\RequirePackage{fix-cm}
\documentclass{svjour3}                     % onecolumn (standard format)
\smartqed  % flush right qed marks, e.g. at end of proof
\usepackage{graphicx}
\usepackage{scrtime}
\usepackage[dont-mess-around]{fnpct}

\usepackage{etoolbox}
\newcommand*{\affaddr}[1]{#1} % No op here. Customize it for different styles.
\newcommand*{\affmark}[1][*]{\textsuperscript{#1}}

\usepackage{url}

\usepackage{breakurl}
\usepackage[breaklinks]{hyperref}

\usepackage{natbib}
\usepackage{caption}
\usepackage{enumitem} % allows costimizing itemsep
\setlist[1]{itemsep=2pt} % sets spacing for first level items 
\setlist[2]{itemsep=1pt} % sets spacing for second level items 

\usepackage{mathtools}
\usepackage{amssymb}
\usepackage[ruled,vlined,linesnumbered]{algorithm2e}
\SetCommentSty{mycommfont}

\SetKwFor{RepTimes}{repeat}{times}{end}
\SetNlSty{bfseries}{\color{black}}{}
\usepackage[dvipsnames]{xcolor}
\DontPrintSemicolon

\SetCommentSty{mycommfont}
\newlength{\arrow}

\newcommand{\FPath}{\mathit{F\mkern-5muP\mkern-4muath}}
\newcommand{\IPath}{\mathit{I\mkern-5muP\mkern-4muath}}
\newcommand{\Path}{\mathit{P\mkern-4muath}}

\newcommand{\unifier}{\mathit{unifier}}
\newcommand{\true}{\mathit{true}}
\newcommand{\false}{\mathit{false}}
\newcommand{\adom}{\mathit{adom}}
\usepackage{multicol}
\usepackage{pgfplots}
\usepackage{tikz}
\usetikzlibrary{shapes}
\newcommand*\circled[1]{\tikz[baseline=(char.base)]{
            \node[shape=circle,draw,inner sep=2pt] (char) {#1};}}

\usepackage{graphicx}
\newtheorem{define}{Definition}
\usepackage{hyperref}
\newcommand{\doi}[1]{\textsc{doi}: \href{http://dx.doi.org/#1}{\nolinkurl{#1}}}
\usepackage{textgreek}
\usepackage{floatrow}
\mathchardef\mhyphen="2D
\let\oldnl\nl% Store \nl in \oldnl
\newcommand{\nonl}{\renewcommand{\nl}{\let\nl\oldnl}}% Remove line number for one line

\begin{document}

\title{Lifted Model Checking for Relational MDPs
% \thanks{Grants or other notes
% about the article that should go on the front page should be
% placed here. General acknowledgments should be placed at the end of the article.}
}
% \subtitle{Do you have a subtitle?\\ If so, write it here}

% \titlerunning{Lifted Model Checking for Relational MDPs}        % if too long for running head

\author{
\begin{tabular}{rl}
Wen-Chi Yang\affmark[1] & \url{wenchi.yang@cs.kuleuven.be} \cr
Jean-François Raskin\affmark[2] & \url{jean-francois.raskin@ulb.be} \cr
Luc De Raedt\affmark[1]\affmark[3] & \url{luc.deraedt@cs.kuleuven.be}
\end{tabular}
}

\authorrunning{Wen-Chi Yang, Jean-François Raskin, Luc De Raedt} % if too long for running head

\institute{
    \affaddr{\affmark[1]Department of Computer Science, KU Leuven, Celestijnenlaan 200a - box 2402, 3001 Leuven, Belgium
    }\\
    \affaddr{\affmark[2]Universit\'e libre de Bruxelles - Campus de la Plaine, CP212 - 1050 Bruxelles - Belgium
    }\\
    \affaddr{\affmark[3]Centre for Applied Autonomous Sensor Systems, \"Orebro University, Sweden.
    }
%             \emph{Present address:} of F. Author  %  if needed
}

\date{Received: date / Accepted: date}
% The correct dates will be entered by the editor

\maketitle

\begin{abstract}
Probabilistic model checking has been developed for verifying systems that have stochastic and nondeterministic behavior.
Given a probabilistic system, a probabilistic model checker takes a property and checks whether or not the property holds in that system.
For this reason, probabilistic model checking provide rigorous guarantees.
So far, however, probabilistic model checking has focused on propositional models where a state is represented by a symbol.
On the other hand, it is commonly required to make relational abstractions in planning and reinforcement learning.
Various frameworks handle relational domains, for instance, STRIPS planning and relational Markov Decision Processes.
Using propositional model checking in relational settings requires one to ground the model, which leads to the well known state explosion problem and intractability.
We present pCTL-REBEL, a lifted model checking approach for verifying pCTL properties of relational MDPs.
It extends REBEL, a relational model-based reinforcement learning technique, toward relational pCTL model checking.
PCTL-REBEL is lifted, which means that rather than grounding, the model exploits symmetries to reason about a group of objects as a whole at the relational level.
Theoretically, we show that pCTL model checking is decidable for relational MDPs that have a possibly infinite domain, provided that the states have a bounded size.
Practically, we contribute algorithms and an implementation of lifted relational model checking, and we show that the lifted approach improves the scalability of the model checking approach.

% Insert your abstract here. Include keywords, PACS and mathematical
% subject classification numbers as needed.
\keywords{
% we need 4-6 keywords
model checking \and 
probabilistic computation tree logic (pCTL) \and 
first-order logic \and 
lifted inference \and 
relational MDPs
% \and relational reinforcement learning
% First keyword \and 
% Second keyword \and 
% More
}
\end{abstract}

\section{Introduction}
Probabilistic model checking aims at deciding whether a stochastic model satisfies a given probabilistic property~\citep{Baier:2008,tutorialPRISM}. By doing so, it provides rigorous guarantees about the model.
Markov Decision Processes (MDPs) and Probabilistic Computational Tree Logic (pCTL)~\citep{tutorialPRISM,Prism:2011,Storm:2017}
are standard formalisms for specifying the model and the properties, respectively. 
MDPs are commonly used for modeling sequential decision making problems where the actions have stochastic effects.
PCTL is a temporal logic that expresses model properties over time and allows for probabilistic quantification. A property can be \textit{the machine gives a warning before shutting down with a probability higher than 0.95}, or \textit{the probability is higher than 0.9 that the emergency power supply, after giving a warning, continues to function for at least 10 more minutes}.

It is common in planning and reinforcement learning to make abstraction of the domain elements in order to compactly define models and speed up the computation. However, model checking methods most often operate on explicit-state MDPs~\citep{Baier:2008} thus do not allow for such abstractions. 
This is undesirable since the number of states increases exponentially with the domain size, making it infeasible to explicitly traverse the state space~\citep{slaney2001blocksworld}. 
In such large domains, it is impractical to apply model checking techniques as they lead to a state explosion~\citep{Otterlo:2004}.
For instance, the well known blocks world has 501 states for 5 blocks but over 58 million states for only 10 blocks~\citep{slaney2001blocksworld}. In this paper, we aim at mitigating such state explosions in probabilistic model checking by making relational abstractions and using lifted inference.

\textit{Lifting} is the key to scalablility in relational domains~\citep{kersting2012lifted,van2011lifted,de20071}. It is also central to statistical relational AI (StarAI)~\citep{starai_book}. 
Lifting implies reasoning about a group of objects as a whole at the first-order level, and exploiting the shared relational structures and symmetries in the model. This is done by making abstraction of irrelevant details of the objects.
As an illustration, an object's full identity (e.g. a block's ID number) can be left out as long as it satisfies the property description (e.g. blue). 
There has been a significant interest in such relational representations in reinforcement learning and planning.
For instance,~\citet{dvzeroski2001relational} introduced \textit{Relational Markov Decision Processes (RMDP)}, a first-order generalization of MDPs that succinctly formulates relational models by implicitly defining states in terms of objects and relations~\citep{dvzeroski2001relational,Otterlo:2004}. 
RMDPs have been used in reinforcement learning and planning to compute first-order policies without explicitly constructing the underlying state space~\citep{dvzeroski2001relational,Kersting:2004,wang2008first,logicalMDP2004,Driessens2004,Boutilier:2001,Sanner:2009,yoon2012inductive}. % flat goal-oriented properties 
One especially interesting example is \textit{REBEL}~\citep{Kersting:2004}, the RElational BELlman operator, which we will extend in this paper.
REBEL is a model-based reinforcement learning technique for constructing an optimal policy in a given RMDP. It is also a lifted inference technique that alleviates state explosions.

Motivated by the success of temporal logics and MDPs in probabilistic model checking and in planning, we investigate whether it is possible to lift these approaches to RMDPs.
We show that the answer is positive by introducing \textit{pCTL-REBEL}, a new framework that augments REBEL~\citep{Kersting:2004} with pCTL. 
More specifically, pCTL-REBEL is a relational model checking approach that checks relational pCTL formulae in RMDPs.
In addition to mitigating state explosions, it is important to take one step further to investigate lifted probabilistic model checking for infinite models. 
Although model checking for infinite models is generally undecidable~\citep{Gabbay2003}, a rich body of research has discussed the \textit{state-boundedness} assumption that yields decidable verification of infinite systems~\citep{Hariri2012,Belardinelli2011,Belardinelli2012,DeGiacomo2012,Calvanese2016}. 
These studies almost exclusively focus on non-probabilistic actions. Nevertheless, they provide great insight into relational model checking with pCTL properties. 
We extend the work of~\citet{Belardinelli2012} to the probabilistic setting to obtain the decidability of the model checking problem for a subclass of infinite MDPs.

The key contribution of this paper is twofold. 
First, we introduce a lifted model checking algorithm, pCTL-REBEL, to mitigate the state explosion problem of checking relational MDPs. PCTL-REBEL is fully automated and provides a complete, lifted solution for relational model checking. In order to adapt to the model checking framework, pCTL-REBEL introduces an alternative interpretation of REBEL~\citep{Kersting:2004} in which a state value corresponds to the \textit{probability} that a given formula is satisfied by executions starting from that state.
Second, while model checking is generally undecidable for infinite MDPs, we provide decidability results for a class of infinite MDPs under the state-boundedness condition. In particular, we prove that a finite relational abstraction exists for RMDPs that have an infinitely large domain, and that checking the relational abstraction is equivalent to checking the infinite MDP. This means that strong guarantees for such infinite MDPs can be provided via lifted model checking, as implemented in pCTL-REBEL.

This paper is structured as follows. 
Section~\ref{sec:rmdp definition} provides an overview on basic notions of relational MDPs. 
Section~\ref{sec:model checking on RMDPs} reviews basic notions of model checking and introduces relational pCTL. Section~\ref{sec:problem statement} defines the problem statement of this paper, that is, probabilistic model checking for relational MDPs.
Section~\ref{sec:relational value iteration} defines a relational Bellman update operator (a generalization of REBEL~\citep{Kersting:2004}) for relational value iteration.
Based on Section~\ref{sec:relational value iteration}, Section~\ref{sec:pCTL-REBEL} introduces the main algorithm, pCTL-REBEL, for relational model checking.
Section~\ref{sec:theoretic results} provides theoretic results about obtaining decidability for a subclass of infinite MDPs.
Section~\ref{sec:experiments} reports on experimental evaluation. 
Section~\ref{sec:related work} discusses related work and 
Section~\ref{sec:discussion} highlights the link between relational model checking and safe reinforcement learning. 
Finally, Section~\ref{sec:conclusion} concludes the work.

\section{Relational Notions and Relational MDPs}
\label{sec:rmdp definition} 

This section defines the basic notions of relational MDPs. These notions will be used in the remainder of this paper with the \textit{blocks world} running example. 
We closely follow the notions of the standard first-order logic~\citep{LogicBook:1997} and relational MDPs~\citep{Kersting:2004}.

\subsection{Relational Logic} 

Relational logic generalizes propositional logic with \textit{variables} such that a variable represents a set of constants.
This section provides an overview of relational logic, following the notions of~\citet{LogicBook:1997}.

An \textit{alphabet} is a tuple $\Sigma=\langle R, D\rangle$ where $R$ is a finite set of relation symbols and $D$ is a possibly infinite set of constants. 
Each relation symbol $\mathtt{p} \in R$ has an arity $m\geq 0$. 
An \textit{atom} $\mathtt{p(t_1, ..., t_m)}$ is a relation symbol $\mathtt{p}$ followed by an $m$-tuple of terms $\mathtt{t_i}$. 
A \textit{term} is a variable $\mathtt{W}$ or a constant $\mathtt{c}$. 
A variable (resp. constant) is expressed by a string that starts with an upper (resp. lower) case letter.
A conjunction is a set of atoms, which is implicitly assumed to be \textit{existentially quantified}.
A definite clause $H\leftarrow B$ consists of an atom $H$ and a conjunction $B$, stating that \textit{$H$ is true if $B$ is true}. 
Given an expression $\mathtt{E}$, $vars(\mathtt{E})$ (resp. $consts(\mathtt{E})$, $terms(\mathtt{E})$) denotes the set of all variables (resp. constants, terms) in $\mathtt{E}$.
An expression is called \textit{ground} if it contains no variables.
We shall make the {\bf unique name assumption}, that states all constants  are unequal, that is, $\mathtt{c_1 \not= c_2}$ holds for
different constants $\mathtt{c_1}$ and $\mathtt{c_2}$. A substitution $\theta$ is a set of bindings $\mathtt{\{W_1/t_1, ..., W_n/t_n\}}$ that assigns terms $\mathtt{t_i}$ to variables $\mathtt{W_i}$. A grounding substitution $\theta$ assigns constants to variables in an expression $\mathtt{E}$ such that $\mathtt{E}\theta$ contains no variables.

We shall use the \textbf{Object Identity subsumption} framework  (OI-subsumption)~\citep{Ferilli02}, which means
that any two terms in an expression are unequal, and the pairwise inequalities should be added. 
For instance, under OI-subsumption, the conjunction $\{\mathtt{cl(b), on(X,Y)}\}$ denotes the expression
$\{\mathtt{cl(b)}$, $\mathtt{on(X,Y)}$, $\mathtt{X \neq Y}$, $\mathtt{X \neq b}$, $\mathtt{Y \neq b}\}$.
For ease of writing, when the context is clear, we shall not write these inequalities explicitly. 
A conjunction $\mathtt{A}$ is \textit{OI-subsumed} by conjunction $\mathtt{B}$, denoted by $\mathtt{A}\preceq_{\theta}\mathtt{B}$, if there exists a substitution $\theta$ such that $\mathtt{B}\theta\subseteq \mathtt{A}$. Only substitutions
that satisfy the inequality constraints are allowed.

A \textit{unifier} $\theta$ of two conjunctions $\mathtt{A}$ and $\mathtt{B}$ under OI-subsumption is a substitution such that $\mathtt{A}\theta = \mathtt{B}\theta$. For example, the conjunctions $\{\mathtt{cl(X), on(y, Z)}\}$ and $\{\mathtt{cl(x), on(Y, Z)}\}$ have a unifier $\{\mathtt{X/x,Y/y}\}$.
A \textit{maximally general specialization (mgs)} of two conjunctions $\mathtt{A}$ and $\mathtt{B}$ under OI-subsumption is a conjunction that is OI-subsumed by $\mathtt{A}$ and $\mathtt{B}$, and is not OI-subsumed by any other specializations.
The mgs operation is not always unique under OI-subsumption.
For example, the conjunctions $\{\mathtt{cl(X)}\}$ and $\{\mathtt{on(Y,Z)}\}$ have maximally general specializations $\{\mathtt{cl(X), on(X,Y)}\}$ and $\{\mathtt{cl(X), on(Y,Z)}\}$ that do not OI-subsume one another.

The \textit{Herbrand base} of an alphabet $\Sigma=\langle R, D\rangle$, denoted by $HB^\Sigma$, is the set of all ground atoms that can be constructed from $\Sigma$. 
A \textit{Herbrand interpretation} $s$ is a subset of $HB^\Sigma$ where all atoms in $s$ are $\true$ and all others are $\false$. The set of all Herbrand interpretations determined by $\Sigma$ is denoted by $S^\Sigma$. We shall write $S$ instead of $S^\Sigma$ when the context is clear.
When the domain $D$ is infinite, the set of all Herbrand interpretations $S^\Sigma$ must be infinite.

\subsection{Relational MDP} 
\label{sec:RMDP}

Relational MDPs (RMDPs) generalize explicit-state MDPs in a twofold manner. 
First, RMDPs have structured states. More specifically, an RMDP state is represented by a conjunction of ground atoms whereas an explicit state is represented by a single constant. 
Second, RMDPs allows variables in the state description. In consequence, a set of RMDP states can be represented by one single \textit{abstract state}, which enables reasoning about a set of states as a whole.
In this paper, an RMDP is a variant of the standard RMDP~\citep{Kersting:2004,Boutilier:2001} that varies by allowing the domain to be infinite. This section formally defines RMDPs, following the notions of~\citet{Kersting:2004}.

An RMDP is a tuple $K=\langle \Sigma, \Delta\rangle$ where the alphabet  
$\Sigma=\langle R, D\rangle$ contains a set of relations and a domain, and $\Delta$ is a finite set of abstract transitions.
The alphabet $\Sigma$ determines the state space. A state $s\in S^\Sigma$ is a Herbrand interpretation. 
An \textbf{abstract state} $s'$ is then a conjunction of atoms, representing a set of states, denoted by $s'\Theta = \{s\in S^\Sigma | s \preceq_{\theta} s' \}$. 
\begin{example}
Consider a blocks world with the alphabet $\Sigma=\langle R, D\rangle$ where the relations are $R=\{\mathtt{cl/1, on/2}\}$ and the domain is $D = \{\mathtt{a,b,c}\}$, the abstract state $s=\{\mathtt{cl(A),cl(C), on(A,B)}\}$ represents the following six ground states. 
\begin{align*}
    s\Theta =
\{&\mathtt{\{cl(a){,}cl(c){,}on(a{,}b)\}},
  \mathtt{\{cl(a){,}cl(b){,}on(a{,}c)\}},\\&
  \mathtt{\{cl(b){,}cl(c){,}on(b{,}a)\}},
  \mathtt{\{cl(b){,}cl(a){,}on(b{,}c)\}},\\&
  \mathtt{\{cl(c){,}cl(b){,}on(c{,}a)\}},
  \mathtt{\{cl(c){,}cl(a){,}on(c{,}b)\}}\}
\end{align*}
\end{example}
An \textbf{abstract action} $\alpha \not\in R$ is an atom for an action relation. 
An \textbf{abstract transition} $\delta\in \Delta$ (based on an abstract action $\alpha$) is a finite set of probabilistic transition rules 
$\delta=\{H_1 \xleftarrow{p_1:\alpha} B, ..., H_n \xleftarrow{p_n:\alpha} B\}$ where $B$ (resp. $H_i$) is an abstract state, representing the precondition (resp. postcondition), and $p_i \in [0,1]$ is the transition probability. 
These transition rules $\delta$ denote a proper probability distribution over $H_i$, that is, $\sum^{n}_{i=1} p_i = 1$. 
To ensure that all abstract transitions rely only on information in the current state, we assume that all variables in $H_i$ are also in $B$, that is, $vars(H_i)\subseteq vars(B)$.
\begin{example}
\label{ex: abstract transition}
A blocks world is defined by an RMDP $K=\langle \Sigma, \Delta\rangle$ where $\Sigma=\langle R, D\rangle$, $R=\{\mathtt{cl/1, on/2}\}$ and $\Delta$ contains the following abstract transition $\mathtt{\delta_{move}}$.

\begin{align*}
\delta_{\mathtt{move}}
\left\{\;
\begin{matrix}
    \mathtt{cl(A){,}cl(C){,}on(A,B)
			\xleftarrow{0.9:move(A,B,C)}
			cl(A){,}cl(B){,}on(A,C)}\\
    \mathtt{cl(A){,}cl(B){,}on(A,C)
			\xleftarrow{0.1:move(A,B,C)}
			cl(A){,}cl(B){,}on(A,C)}
\end{matrix}
\right.
\end{align*}

The abstract action $\mathtt{move(A,B,C)}$ expresses \textit{moving block $\mathtt{A}$ to block $\mathtt{B}$ from block $\mathtt{C}$}. The action succeeds with probability $0.9$ and fails with probability $0.1$. When the action fails, the state stays the same. A graphical illustration is in Figure~\ref{fig:abstract transition}.  
\begin{figure}[h]
    \centering
    \captionsetup{justification=centering,margin=2cm}
    \includegraphics[width=0.5\textwidth]{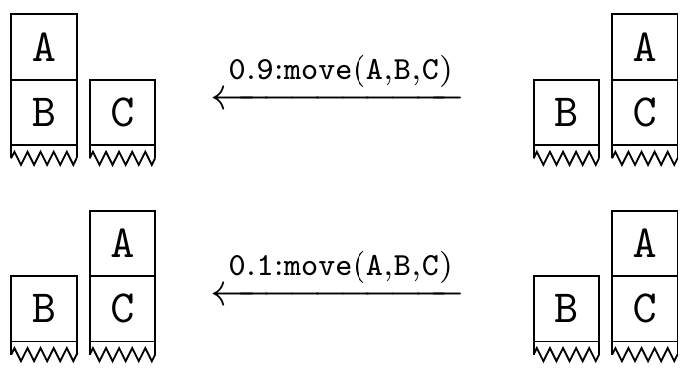}
    \caption{The abstract transition $\mathtt{\delta_{move}}$ moves block $\mathtt{A}$ to block $\mathtt{B}$ from block $\mathtt{C}$ with probability $0.9$. It fails to move the block with probability $0.1$. }
    \label{fig:abstract transition}
\end{figure}
\end{example}

\subsection{Grounding an RMDP}

The semantics of an RMDP is defined at the ground level such that any RMDP (including the infinite ones) implicitly defines an underlying ground MDP. This section formally defines the construction of the underlying ground MDP. 
In the end of this section, we will briefly discuss the decidability issue of infinite RMDPs. 

Given an RMDP $K=\langle \Sigma, \Delta\rangle$, the underlying ground MDP is a tuple $M=\langle S,A,T\rangle$ where $S$ is a set of ground states, $A$ is a set of ground actions and $T:S\times A\times S\rightarrow [0,1]$ is a ground transition function. 
Every \textbf{ground state} $s\in S$ is a Herbrand interpretation of $\Sigma$.
A ground state $s$ has a set of available \textbf{ground actions}, denoted by $A(s)\subseteq A$. These ground actions $A(s)$ are defined by grounding the abstract actions. Formally,
\begin{align*}
A(s) := 
\{
\alpha\theta | H_i \xleftarrow{p_i:\alpha} B \in \Delta, s \preceq_{\theta} B
\} 
\end{align*}
Similarly, given a ground state $s$ and a ground action $\alpha\theta\in A(s)$, the set of \textbf{ground transitions} $T(s,\alpha\theta)$ are defined by grounding the abstract transitions. Formally,
\begin{align*}
T(s,\alpha\theta) := \{ h_i \xleftarrow{p_i:\alpha\theta} s | 
&H_i \xleftarrow{p_i:\alpha} B\in \Delta, s \preceq_{\theta} B,
h_i = (s\backslash B\theta)\cup H_i\theta
\} 
\end{align*}
% The \textbf{ground transition function} of state $s$ and action $a$ is defined for post states $s'$ that can be subsumed by some abstract transition. Formally,
% \begin{align*}
% T(s,\alpha\theta)(s') := p_i 
% \bigg|_{
% \begin{array}{cc}
%      H_i \xleftarrow{p_i:\alpha} B\in \Delta, s \preceq_{\theta} B,\\
%     s' = (s\backslash B\theta)\cup H_i\theta
% \end{array}
% }
% \end{align*}
After taking action $\alpha\theta$ in state $s$, the transition probability distribution over all ground states $s' \in S$ is then
\begin{align*}
    T(s, \alpha\theta)(s') := p_i, \mbox{ where } s'=(s\backslash B\theta)\cup H_i\theta
\end{align*}
Since $\Delta$ is a proper abstract transition function, $T$ must be a proper probability distribution, i.e. $\sum_{s'\in S} T(s,\alpha\theta)(s') = 1$. 

\begin{example}
Consider a blocks world defined by an RMDP $K=\langle \Sigma, \Delta\rangle$ where $\Sigma=\langle R, D\rangle$, $R=\{\mathtt{cl/1, on/2}\}$, $D=\{\mathtt{a,b,c,d,e}\}$ and $\Delta=\{\mathtt{\delta_{move}}\}$ (see Figure~\ref{fig:abstract transition}).
The RMDP $K$ defines the underlying MDP $M=\langle S, A, T\rangle$. One of the resulting ground transitions is as follows.
Let a ground state be $s = \{\mathtt{cl(a)}$, $\mathtt{cl(b)}$, $\mathtt{cl(d)}$, $\mathtt{on(a,c)}$, $\mathtt{on(d,e)}\} \in S$, and let an action be $\mathtt{move(a,b,c)} \in A(s)$,
the resulting next state must be $s' = \{\mathtt{cl(a)}$, $\mathtt{cl(c)}$, $\mathtt{cl(d)}$, $\mathtt{on(a,b)}$, $\mathtt{on(d,e)}\}\in S$. By taking the $\mathtt{move(a,b,c)}$ action in state $s$, the probability of reaching $s'$ is 0.9, and the probability of staying in $s$ is 0.1, as illustrated in Figure~\ref{fig:ground transition}. 
\begin{figure}[h]
    \centering
    \captionsetup{justification=centering,margin=2cm}
    \includegraphics[width=0.5\textwidth]{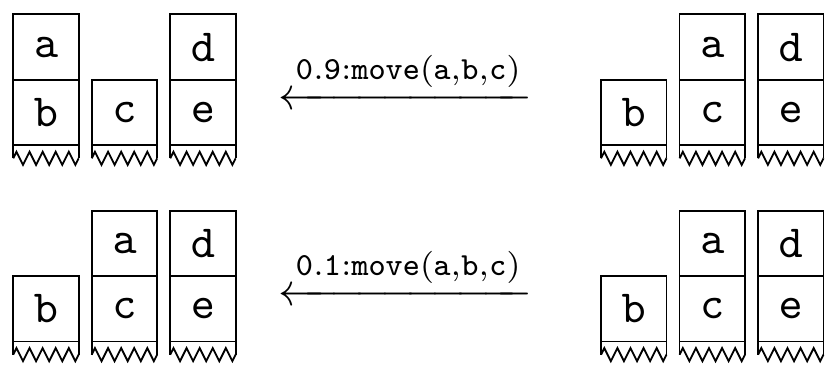}
    \caption{The ground action $\mathtt{move(a,b,c)}$ moves block $\mathtt{a}$ to block $\mathtt{b}$ from block $\mathtt{c}$ with probability $0.9$. The action fails with probability $0.1$. }
    \label{fig:ground transition}
\end{figure}
\end{example}

In general, for an RMDP that has an infinite domain, the underlying ground MDP has an infinitely large state space and action space. Hence, it is infeasible to explicitly traverse the state space.
Moreover, such RMDPs have an unbounded branching behavior such that a state has infinitely many available actions, leading to infinitely many other states. For example, one can move a clear block to any of the (infinitely many) other clear blocks. 
Therefore, the model checking problem for infinite RMDPs is generally undecidable. 
To obtain decidability, we will later identify a special class of infinite RMDPs such that the branching behavior is bounded.
More details are in Section~\ref{sec:theoretic results}.

\section{Model Checking for Relational MDPs}
\label{sec:model checking on RMDPs}
This section defines the problem statement of this paper, namely the \textit{model checking problem of relational MDPs}.
More specifically, Section~\ref{sec:reachability} reviews the fundamentals of model checking, Section~\ref{sec:pCTL} defines the relational pCTL language that will be used to specify properties throughout this paper, and Section~\ref{sec:problem statement} defines the main problem statement. 
Later, Section~\ref{sec:relational value iteration} and Section~\ref{sec:pCTL-REBEL} will provide a solution for the problem in Section~\ref{sec:problem statement}.

\subsection{Probabilistic Reachability}
\label{sec:reachability}

This section defines the \textit{probabilistic reachability property}, the most fundamental property in model checking. 
The probabilistic reachability refers to the maximum probability of reaching a set of goal states from a given initial state. 
This property is required to define the relational pCTL language in Section~\ref{sec:pCTL}. 
The content of this section is standard in model checking~\citep{Baier:2008,Prism:2011} and is included to make this paper self-contained.

We first define \textit{paths} in an MDP and their probability assignment. 
Given an MDP $M=\langle S,A,T\rangle$, a path of length $n$ is denoted by $\rho_n = s_1\xrightarrow{a_1} s_2\xrightarrow{a_2} ... \xrightarrow{a_{n-1}} s_n$ where $s_i\in S$, $a_i\in A(s_i)$ and $T(s_i,a_i)(s_{i+1}) > 0$.
Similarly, an infinite path is denoted by $\rho = s_1\xrightarrow{a_1} s_2\xrightarrow{a_2} ... $. 
The set of all finite (resp. infinite) paths is denoted by $\FPath_{M}$ (resp. $\IPath_{M}$), and
the set of all finite (resp. infinite) paths starting from state $s$ is denoted by $\FPath_{M,s}$ (resp. $\IPath_{M,s}$). 
We denote all paths starting from $s$ in $M$ as $\Path_{M,s} {=} \FPath_{M,s}\cup\IPath_{M,s}$.
The i-th state of a path $\rho$ is denoted by $\rho(i)$. The last state of a path $\rho$ is denoted by $last(\rho)$. To project a set of paths $\Path_{M,s}$ to a probability space, it is required to remove nondeterministic actions by a policy. 
A policy $\pi: \FPath_{M}\times A\rightarrow [0,1]$ takes a finite path $\rho_n$ and specifies a probability distribution over all available actions $A(last(\rho_n))$. 
We write $\pi(\rho_n)$ to denote all possible actions selected by the policy, and we write $\pi(\rho_n, a)$ to denote the probability of action $a$ being selected.

\begin{define}
\label{def:prob measure for FPath}
(cf.~\citet{tutorialPRISM})
Given an MDP $M$ and a policy $\pi$, the probability $P_{M}^{\pi}(\rho_n)$ of a finite path $\rho_n = s_1\xrightarrow{a_1} s_2\xrightarrow{a_2} ... \xrightarrow{a_{n-1}} s_n \in \FPath_{M,s_1}$ is inductively defined by 
\begin{align}
&P_{M}^{\pi}(\rho_{1}) = 1 \nonumber\\
\label{eq:path probability function}
&P_{M}^{\pi}(\rho_{k}) = 
P_{M}^{\pi}(\rho_{k-1}) 
\sum_{a\in \pi(\rho_{k-1})}
\pi(\rho_{k-1},a) T(s_{k-1}, a)(s_{k})
\end{align}
where $\rho_k$ denotes $\rho_n$'s prefix of length $k$. 
% Since the starting state $s_1$ is implicitly defined in $\rho$, we write $P_{M}^{\pi}(\rho)$ instead of $P_{M,s_1}^{\pi}(\rho)$ from now on.
Let $C_{\rho_n}\in\IPath_{M,s_1}$ be the set of all infinite paths that have a prefix $\rho_n$ (also known as the basic cylinder), the probability $P^\pi_{M}(C_{\rho_n})$ is then defined as the probability of $\rho_n$, i.e. $P^\pi_{M}(C_{\rho_n}) = P^\pi_{M}(\rho_n)$.
\end{define}

After defining the probability for an MDP path, we can now formally define probabilistic reachability.

\begin{define} (cf.~\citet{tutorialPRISM})
Given an MDP $M=\langle S, A, T\rangle$, an initial state $s\in S$ and a set of goal states $G \subseteq S$, the set of all paths from $s$ to $G$, denoted by $\Path_{M,s}(G)$, is formally defined as follows. 
\begin{align*}
\Path_{M,s}(G) := \{ \rho\in\Path_{M,s} |\exists i \in \mathbb{N}, \rho(i) \in G \}
\end{align*}
\end{define}
By following an optimal policy $\pi$, one can generate a path from $s$ to $G$ with a maximum probability. 
Any other policies in the policy space $\Pi$ does not achieve a probability larger than the optimal policy $\pi$ does. The maximum reachability is defined as follows.
\begin{align}
\label{eq: max probability}
P^{max}_{M} (\Path_{M,s}(G)) &:= \sup_{\pi\in\Pi} P^{\pi}_{M}(\Path_{M,s}(G))
\end{align}
In general, $P^{max}_{M}(.)$ denotes the maximum reachability in $M$. 

It is known that a deterministic, stationary $\epsilon$-optimal policy suffices to achieve the maximum probabilistic reachability in a finite MDP~\citep{Baier:2008}.
This means the policy will simplify to a function $\pi: S\rightarrow A$ that maps a path's last state to a single action. For an infinite RMDP that has a finite abstraction (cf. Section~\ref{sec:theoretic results}), we assume there exists an optimal deterministic, stationary policy for the maximum probabilistic reachability.

\subsection{Relational pCTL} 
\label{sec:pCTL}

This section introduces \textit{Relational Probabilistic Computational Tree Logic} (relational pCTL), a temporal logic that describes system behavior over time and allows for probabilistic quantification. We will use relational pCTL to specify properties of an RMDP. Relational pCTL is a variant of the standard pCTL (cf.~\citet{tutorialPRISM} and~\citet{Baier:2008}) that varies by allowing variables in atoms and only allowing negations in front of atoms. 

The syntax of the relational pCTL is as follows. A property is always specified by a state formula $\phi$.
\begin{align*}
state\; formula\;\phi &::= \mathtt{true} \;|\; l \;|\; \neg l \;|\; \phi \land \phi \;|\; \phi \lor \phi \;|\; \mathtt{P_{\bowtie p}} [\psi] \\
path\; formula\;\psi &::= \mathtt{X} \;\phi \;|\; \phi\;\mathtt{U^{\leq k}}\; \phi \;| \; \phi\;\mathtt{U}\; \phi
\end{align*}
where $l$ is an atom (that can contain variables), $\mathtt{p}$ is a probability such that $0 \leq \mathtt{p} \leq 1$, $\mathtt{k} \in \mathbb{N}$ is a step bound and $\bowtie\; \in \{\leq, <, \geq, >\}$. 
Here, relational pCTL generalizes the standard pCTL by letting $l$ be a relational atom instead of a constant.

The semantics of the relational pCTL resembles the standard pCTL~\citep{Baier:2008}. A state either satisfies or violates a state formula $\phi$, resulting in a boolean evaluation for each state. The $\mathtt{X}$ operator stands for \textit{next}, and the $\mathtt{U}$ stands for \textit{until}. A path formula $\mathtt{X}\phi$ is satisfied if $\phi$ is satisfied in the next state;  $\phi_1 \mathtt{U^{\leq k}} \phi_2$ is satisfied if $\phi_2$ is satisfied within $\mathtt{k}$ steps and $\phi_1$ holds before then;  $\phi_1 \mathtt{U} \phi_2$ is satisfied if $\phi_2$ is eventually satisfied and $\phi_1$ holds before then.

The semantics of the relational pCTL is defined at the ground level. Given an RMDP $K=\langle\Sigma,\Delta\rangle$ that defines a ground MDP $M=\langle S, A, T\rangle$, we say a ground state $s\in S$ satisfies a state formula $\phi$, denoted by $ s\models \phi$, if and only if there exists a grounding substitution $\theta$ for all free variables in $\phi$ such that $s$ satisfies $\phi$ under $\theta$, i.e.
$
s\models \phi \Leftrightarrow \exists \theta . s\models^\theta \phi
$. All substitutions must respect OI-subsumption, i.e. any two terms $\mathtt{t_1, t_2}$ in a conjunction must be unequal. 
% Similarly, $s$ satisfies a path formula $\psi$, denoted by $s\models \psi$, if and only if there exists a grounding substitution $\theta$ for all free variables in $\psi$ such that $s$ satisfies $\psi$ under $\theta$, i.e.
% $
%     s\models \psi \Leftrightarrow \exists \theta . s\models^\theta \psi
% $
Formally, the pCTL satisfiability relation $\models^\theta$ is inductively defined as follows. 
\begin{align*}
s &\models^\theta \mathtt{true} &&\\
s &\models^\theta l &&\Leftrightarrow s \preceq_{\theta} l\\
s &\models^\theta \neg l &&\Leftrightarrow s \not\preceq_{\theta} l\\
s &\models^\theta \phi_1 \land \phi_2 &&\Leftrightarrow s\models^\theta \phi_1 \land s\models^\theta \phi_2\\
s &\models^\theta \phi_1 \lor \phi_2 &&\Leftrightarrow s\models^\theta \phi_1 \lor s\models^\theta \phi_2\\
% s &\models^\theta \neg \phi &&\Leftrightarrow s \not\models^\theta \phi\\
s &\models^\theta \mathtt{P_{\bowtie p}} [\psi] &
% \Leftrightarrow s\models \mathtt{P}_{\bowtie p} [\psi]
&\Leftrightarrow  P_{M}^{max}(\{\rho \in \Path_{M,s} | \rho \models \psi\}) \bowtie \mathtt{p}
\end{align*}
where
\begin{align*}
\rho &\models \mathtt{X} \;\phi 
&&\Leftrightarrow \rho(2) \models \phi \\ 
\rho &\models \phi_1 \; \mathtt{U^{\leq k}} \; \phi_2
&&\Leftrightarrow \exists i {\leq} k+1. [ \rho(i)\models \phi_2 \land \forall j{<}i. \rho(j)\models \phi_1 ]\\ 
\rho &\models \phi_1 \; \mathtt{U} \; \phi_2
&&\Leftrightarrow \exists i {\in} \mathbb{N}. [ \rho(i)\models \phi_2 \land \forall j{<}i. \rho(j)\models \phi_1 ]\\
% G
% \rho &\models \mathtt{G^{\leq k}} \; \phi
% &&\Leftrightarrow \forall i {\leq} k. \rho(i)\models \phi \\
% \rho &\models \mathtt{G} \; \phi
% &&\Leftrightarrow \forall i {\in} \mathbb{N}. \rho(i)\models \phi \\
\end{align*}

With the aforementioned operators, additional operators can be defined as follows where $\mathtt{F}$ stands for \textit{eventually}.

\begin{multicols}{3}
\begin{itemize}[label={}]
\item $\mathtt{false} \equiv \neg \mathtt{true}$
% \item $\phi_1 \vee \phi_2 \equiv \neg (\neg\phi_1 \wedge \neg\phi_2)$
% \item $\phi_1 \rightarrow \phi_2 \equiv \neg\phi_1 \vee \phi_2$
\item $\mathtt{F^{\leq k}} \;\phi  \equiv \mathtt{true \;U^{\leq k}}\; \phi$
\item $\mathtt{F} \;\phi  \equiv \mathtt{true \;U}\; \phi$
% \item $\mathtt{P}_{\triangleright p}[\mathtt{G}\;\phi] \equiv \mathtt{P}_{\triangleleft 1-p}[\mathtt{F}\;\neg\phi]$
\end{itemize}
\end{multicols}

\begin{example}
Consider a formula $\phi=\neg \mathtt{cl(A)}$ that states "there exists an unclear block $\mathtt{A}$ in the state". The satisfiability relation $s\models \phi$ can be rewritten as follows. 
\begin{align*}
    &s \models \neg \mathtt{cl(A)}\\
\Leftrightarrow\;& 
    \exists\theta .s \models^\theta \neg \mathtt{cl(A)}\\
% \Leftrightarrow\;&  
%     \exists\theta .s \not\models^\theta \mathtt{cl(A)}\\
\Leftrightarrow\;&  
    \exists\theta .s\not\preceq_{\theta} \mathtt{cl(A)}
\end{align*}
For the ground state $s_1 = \{\mathtt{cl(a)}\}$,  the above evaluates to $\false$ as no $\theta$ exists for the satisfiability relation. For another ground state $s_2 = \{\mathtt{cl(a), on(a,b)}\}$, the above evaluates to $\true$ with $\theta=\{\mathtt{A/b}\}$.
\end{example}

\begin{example}
Consider a formula $\phi=\mathtt{P}_{\geq 0.9} [\mathtt{X\; cl(A)}]$ that states "there exists a block $\mathtt{A}$ that can become clear in the next state with a probability $\geq$ 0.9". 
The satisfiability relation $s\models \phi$ can be rewritten as follows.
\begin{align*}
    &s \models \mathtt{P}_{\geq 0.9} [\mathtt{X\; cl(A)}]\\
\Leftrightarrow\;& 
    \exists\theta . s \models^\theta \mathtt{P}_{\geq 0.9} [\mathtt{X\; cl(A)}]\\
\Leftrightarrow\;& 
    \exists\theta . s \models^\theta  P^{max}_M(\{\rho\in\Path_{M,s}|\rho(2)\models\mathtt{cl(A)}\}) \geq 0.9
\end{align*}
For the ground state $s=\{\mathtt{cl(a)}$, $\mathtt{on(a,b)}\}$, the above evaluates to $\true$ with $\theta = \{\mathtt{A/b}\}$ as block $\mathtt{b}$ can be clear after taking the action $\mathtt{move(a,fl,b)}$ where $\mathtt{fl}$ stands for floor. 
\end{example}

It is assumed that the scope of OI-subsumption is within the conjunction. That is, no term inequalities are assumed across different conjunctions. For example, the formula $\mathtt{P}_{\geq 0.7}[\mathtt{X\; cl(A)}]\land\mathtt{P}_{\geq 0.95}[\mathtt{F\; on(B,C)}]$ has two conjunctions, and term inequalities such as $\mathtt{A = B}$ or $\mathtt{A \neq C}$ do not exist, but $\mathtt{B\neq C}$ holds under OI-subsumption.

\subsection{The Relational Model Checking Problem}
\label{sec:problem statement}

This section defines the model checking problem for RMDPs, using the definition of RMDP (Section~\ref{sec:RMDP}) and relational pCTL (Section~\ref{sec:pCTL}). Later, Section~\ref{sec:relational value iteration} and Section~\ref{sec:pCTL-REBEL} will illustrate techniques for solving this model checking problem. 

Relational model checking resembles the standard model checking problem~\citep{Baier:2008}. Given a model $M$ and a pCTL formula $\phi$, relational model checking computes \textit{all states in $M$ that satisfy $\phi$}, denoted by $Sat_M(\phi)$. The significance of relational model checking is that it computes $Sat_M(\phi)$ at a lifted level by using relational states to represent groups of underlying ground states. Hence, relational model checking finds a set of \textit{abstract states} that represents $Sat_M(\phi)$. 

\begin{define}
\label{Problem definition}
Given an RMDP $K=\langle\Sigma,\Delta\rangle$ that defines the underlying MDP $M=\langle S,A,T\rangle$ and a relational pCTL formula $\phi$, the relational model checking problem is to determine all ground states $Sat_M(\phi)\subseteq S$ that satisfy $\phi$, i.e. $Sat_M(\phi) = \{s\in S | s\models\phi\}$. 
It does so by finding a set of abstract states $Sat_K(\phi)$ in $K$ that represents $Sat_M(\phi)$. Formally,
$$ 
s \in Sat_M(\phi) \Leftrightarrow \exists  s' \in Sat_K(\phi).  s\preceq_{\theta} s'
$$
\end{define}

In this work, we will solve the relational model checking problem of two types of RMDPs. The first type is the RMDPs that have a finite domain (i.e. finite RMDPs). The second type is a special class of RMDPs that have an infinite domain (i.e. infinite RMDPs). 

We discuss the decidability of these two types of RMDPs.
The model checking problem for a finite RMDP is decidable. This is because when the domain is finite, the state space is also finite, i.e. the underlying ground MDP is finite. One can thus enumerate all states and collect the states that satisfy the given property. In contrast, an infinite RMDP contains infinitely many states, which makes enumerating all states infeasible. In this case, we focus on a class of infinite RMDPs that have a finite abstraction. More details are given in Section~\ref{sec:theoretic results}.

\section{PCTL Relational Bellman Operator} 
\label{sec:relational value iteration}

This section defines the \textbf{pCTL relational Bellman operator} (pCTL-REBEL), the essential building block for solving the relational model checking problem. 
Given a pCTL formula $\mathtt{P_{\bowtie p}[\psi]}$ and an RMDP, pCTL-REBEL evaluates a function $V^p: S\rightarrow [0,1]$ that assigns a probability to each RMDP state. A probability $V^p(s)$ represents
the probability that state $s$ satisfies the path formula $\psi$. If the probability is within the bound, i.e. $V^p(s)\bowtie \mathtt{p}$, then state $s$ satisfies $\mathtt{P_{\bowtie p}[\psi]}$ and belongs to the solution set, i.e. $s\in Sat_K(\mathtt{P_{\bowtie p}[\psi]})$.

At this point, it is important to remark that 
PCTL-REBEL is a variant of REBEL, but does not consider a reward structure. Indeed,
REBEL~\citep{Kersting:2004} is a model-based relational reinforcement learning technique that operates on a \textit{reward structure} and computes an optimal policy for reaching a set of goal states, which can be seen as a \textit{reward-based} reachability property. However, since we do not consider a reward structure and are interested in \textit{probabilistic} properties, an alternative interpretation of REBEL is required.
Section~\ref{sec:relational Bellman operator} introduces an alternative interpretation of the relational Bellman operator. Based on which, Section~\ref{sec:logical regression}, \ref{sec:Q rules} and \ref{sec:V rules} respectively describe in detail the three components of pCTL-REBEL. Section~\ref{sec: illustration} then gives an illustration of pCTL-REBEL with an example.

\subsection{PCTL Relational Bellman Operator}
\label{sec:relational Bellman operator}

Given an RMDP and a pCTL formula $\mathtt{P_{\bowtie p}}[\psi]$, the task of pCTL-REBEL is to compute a \textit{state probability function} $V^p: S\rightarrow [0,1]$ that assigns a probability to each state. Similar to the original REBEL, pCTL-REBEL takes an initial state probability function $V^p_0$ and iteratively computes $V^p_1, V^p_2$, etc for a number of steps, depending on the given formula. When the formula has a \textit{step bound} $\mathtt{k}$, then Then pCTL-REBEL is applied for $\mathtt{k}$ times \footnote{A step bound is commonly called a finite horizon in AI.}. When the formula is unbounded, pCTL-REBEL is applied for arbitrarily many times until the probabilities converge.

The state probability function $V^p$ is similar to REBEL's state value function $V$ but interprets state values as probabilities rather than as expected rewards. 
In a similar way, the state-action probability function $Q^p$ is related to REBEL's $Q$ function but interprets state-action values as probabilities. More details will be given later. In order to maintain the connection to the original REBEL and to leave room for extending the present model checking approach to incorporate rewards, we use the same notation $V$ and $Q$ for these functions as in REBEL. For clarification, we add a superscript $p$ to denote that $V^p$ and $Q^p$ interpret values as probabilities. 
We now formally define these functions and the pCTL relational Bellman operator, following the notations of~\citet{Kersting:2004}.

\begin{define} (cf.~\citet{Kersting:2004})
A state probability function $V^p: S\rightarrow [0,1]$ is an ordered set of $V^p$-rules of the form of $c \leftarrow B$ where $B$ is an abstract state and $c \in [0,1]$, representing the probability of reaching a goal state from $B$. The value $V^p(s)$ of a ground state $s$ is assigned by the first rule that subsumes $s$, i.e. $s\preceq_\theta B$. 
\end{define}
Given an abstract goal state $G$, the initial state probability function $V^p_0$ is defined as
\begin{align*}
	1.0 &\leftarrow G \\
	0 &\leftarrow \emptyset
\end{align*}

The first rule expresses that any ground state subsumed by the goal state $G$, by definition, satisfies $G$ with probability 1. The second rule expresses that any other states that are not captured by the first rule satisfy G with probability 0. The rule of $0\leftarrow\emptyset$ ensures that all states are assigned a value. Hence, it is often the last $V^p$-rule to capture the states that are not captured by any previous rules.

\begin{define} (cf.~\citet{Kersting:2004})
A state-action probability function $Q^p:S\times A\rightarrow [0, 1]$ is an ordered set of $Q^p$-rules of the form of $c: A \leftarrow B$ where $A$ is an abstract action and $B$ is an abstract state, representing the probability of reaching a goal state when $A$ is taken in $B$. The value $Q^p(s,a)$ of a ground state $s$ and an action $a$ is assigned by the first rule that subsumes $s$ and $a$, i.e. $s\preceq_\theta B$ and $a\preceq_\theta A$.
\end{define}

The pCTL-REBEL operator is listed in Equation~\ref{eq:pCTL REBEL update}. By iteratively applying pCTL-REBEL, we  compute
the state probability functions $V^p_1, V^p_2$, etc. 
Notice that pCTL-REBEL is a special case of the original REBEL that sets the discount factor to 1, has no reward structure, and connects a single reward 1 to the target condition \footnote{Our work addresses a special case of relational model-based reinforcement learning. More details will be given in Section~\ref{sec:discussion}.}.  As a result, all $V^p$ values are interpreted as probabilities in [0,1]. This alternative interpretation allows to capture the probability that a formula is satisfied, which is essential for adapting the original REBEL framework into a model checking setting.

\begin{align}
\label{eq:pCTL REBEL update}
V^p_{t+1}\underbrace{(s)}_{\text{\circled{1}}} &= 
\overbrace{
\max_{a\in A(s)} 
\underbrace{
\sum_{s'} 
T(s,a,s')V^p_{t}(s')
}_{\text{\circled{2}}}
}
^{\text{\circled{3}}}
= \max_{a\in A(s)} {Q^p}_{t+1}(s,a) 
\end{align}

PCTL-REBEL (Equation~\ref{eq:pCTL REBEL update}) is implemented by
\texttt{OneIteration} (Algorithm~\ref{algo:OneIteration}). This algorithm makes use of the following three components.
\begin{itemize}
    \item [\textcircled{\small{1}}] \texttt{Regression} (Algorithm~\ref{algo:Regression}): Deriving the abstract states $s$ in $V^p_{t+1}$ 
    \item [\textcircled{\small{2}}] \texttt{$Q^p$Rules} (Algorithm~\ref{algo:QRules}): Computing $Q^p_{t+1}(s,a)$ for all actions $a\in A(s)$ 
    \item [\textcircled{\small{3}}] \texttt{$V^p$Rules} (Algorithm~\ref{algo:VRules}): Updating $V^p_{t+1}$ by maximizing over $Q^p_{t+1}(s,a)$. 
\end{itemize}
These three components modify the algorithms of~\citet{Kersting:2004} to provide support to pCTL operators. The modified parts in the algorithms will be marked \textcolor{blue}{blue}. 
We now describe these components in detail in Section~\ref{sec:logical regression}, \ref{sec:Q rules} and \ref{sec:V rules}, respectively.

\begin{algorithm}[ht]
    \SetKwInOut{Input}{Require}
    \SetKwInOut{Output}{Return}
    \Input{
    \begin{tabular}{ c l }
        $V^p_t$ &: state probability function \\
        $\psi$ &: pCTL formula $[\mathtt{S_1\;U\;S_2}]$ or $[\mathtt{X\; S_2}]$
    \end{tabular}
    }
    \Output{
    \begin{tabular}{ c l }
        $V^p_{t+1}$ &: the next probability function
    \end{tabular}    
    }
    \nonl\hrulefill\\
	${Q^p}_{t+1}:=$ \texttt{$Q^p$Rules}$(V^p_t, \textcolor{blue}{\mathtt{S_1}})$\;
	$V^p_{t+1}:=$ \texttt{$V^p$Rules}$({Q^p}_{t+1}, \textcolor{blue}{\psi})$\;
	\caption{
	\texttt{OneIteration}. This algorithm implements Equation~\ref{eq:pCTL REBEL update}.  }
	\label{algo:OneIteration}
\end{algorithm}

\subsection{Logical Regression}
\label{sec:logical regression}

\textit{Logical regression} is a standard technique that reasons about abstract transitions at the relational level~\citep{Boutilier:2001,Kersting:2004,Sanner:2009}.  
This technique is essential to scalability as it mitigates state explosions by operating on the relational state space instead of the underlying ground state space.
Concretely, logical regression searches \textit{backwards} the possible \textit{pre-states} that can reach a given state after taking a number of transitions.

This section extends the standard logical regression to implement Equation~\ref{eq:pCTL REBEL update} \textcircled{\small{1}}.
The task is to identify the pre-states that can reach a given state by taking \textit{one transition}, which is related to computing pCTL formulae with a step bound 1. These formulae are
of the form of $[\mathtt{X\; S_2}]$, $[\mathtt{F^{\leq 1}S_2}]$ or $[\mathtt{S_1\;U^{\leq 1}S_2}]$. 
These three formulae are different in terms of constraints. 
% The path operators $\mathtt{U^{\leq 1}}$ and $\mathtt{X}$ are actually a $\mathtt{F^{\leq 1}}$ operator with additional constraints. 
First, $[\mathtt{F^{\leq 1}S_2}]$ is a {reachability property} that does not impose any constraints. This formula is the basis for the other two formulae.
Second, $[\mathtt{X}\; \mathtt{S_2}]$ imposes the constraint that "$\mathtt{S_2}$ must be reached after taking \textit{exactly one transition}". 
Finally, $[\mathtt{S_1}\;\mathtt{U^{\leq 1}}\mathtt{S_2}]$ is a \textit{constrained reachability property} that imposes the constraint that "$\mathtt{S_2}$ must be reached by going through \textit{only the states where $\mathtt{S_1}$ holds}".

% We first illustrate logical regression with the blocks world example, then discuss its implementation \texttt{Regression} (Algorithm~\ref{algo:Regression}).

\begin{example}
To illustrate logical regression, consider an abstract state $\{\mathtt{on(a,b)}\}$ and the following $\delta_{\mathtt{move_1}}$ transition rule (also shown in Figure~\ref{fig:abstract transition}).
\begin{align*}
\delta_{\mathtt{move_1}}:
    \mathtt{cl(A){,}cl(C){,}on(A,B)
			\xleftarrow{0.9:move(A,B,C)}
			cl(A){,}cl(B){,}on(A,C)}
\end{align*}
By applying the substitution $\theta=\{\mathtt{A/a, B/b}\}$ to $\delta_{\mathtt{move_1}}$, we obtain the following rule. This rules describes that $\{\mathtt{on(a,b)}\}$ can be reached from any ground state that is subsumed by $\mathtt{\{cl(a), cl(b),on(a, C)\}}$ after taking the action $\mathtt{move(a,b,C)}$.
\begin{align*}
    \mathtt{on(a,b)\xleftarrow{0.9:move(a,b,C)} cl(a), cl(b),on(a, C)}
\end{align*}
\end{example}

\begin{figure}[h]
\centering
\includegraphics[width=.7\textwidth]{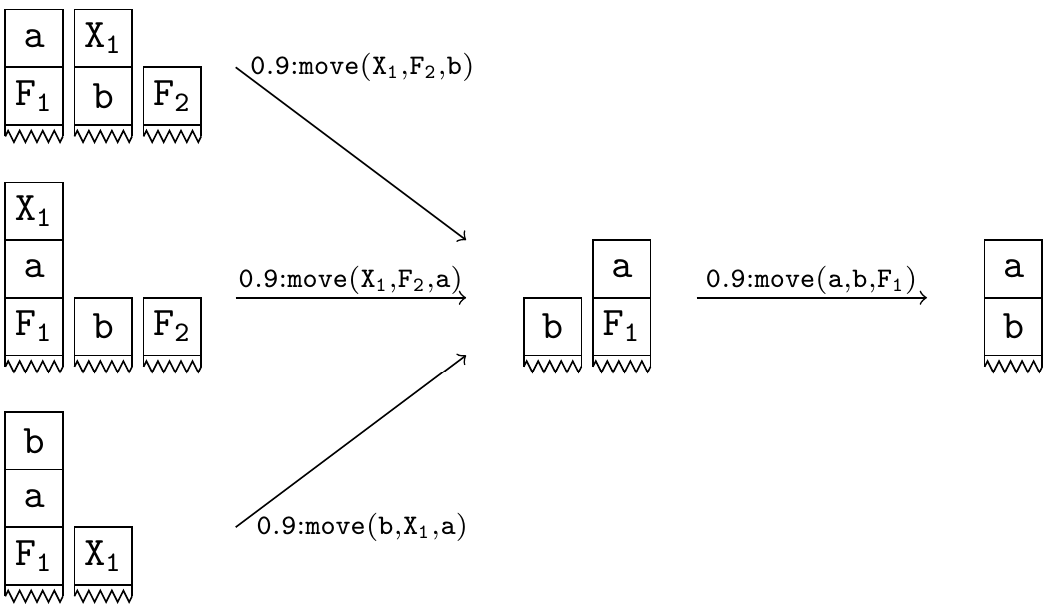}
\captionsetup{width=.7\linewidth}
\caption[Caption for LOF]{
The figure shows the derived paths by applying two logical regressions on $\{\mathtt{on(a,b)}\}$ with the $\delta_{move_1}$ transition rule. 
The abstract states from left to right reach $\{\mathtt{on(a,b)}\}$ after 2, 1, and 0 steps respectively. 
} \label{fig:path}
\end{figure}

Logical regression can be applied multiple times. For example, by applying twice the $\delta_{\mathtt{move_1}}$ transition rule, we obtain all abstract states that can reach $\{\mathtt{on(a,b)}\}$ within 2 steps, as shown in Figure~\ref{fig:path}.

Logical regression is implemented in \texttt{Regression} (Algorithm~\ref{algo:Regression}) that identifies all states that can reach a goal state after taking one transition. 
\texttt{Regression} generalizes the \texttt{WEAKESTPRE} algorithm in REBEL~\citep{Kersting:2004} in two ways (marked \textcolor{blue}{blue}).
First, it provides support to \textit{until} formulae $[\mathtt{S_1\;U\;S_2}]$ by ensuring that all pre-states satisfy $\mathtt{S_1}$. 
Second, it provides support to infinite RMDPs by filtering out the pre-states that exceed a given state bound $b$. More details will be given in Section~\ref{sec: pCTL REBEL infinite MDPs}.

\begin{algorithm}[ht]
    \SetKwInOut{Input}{Require}
    \SetKwInOut{Output}{Return}
    \Input{
    \begin{tabular}{ c l }
        $\delta$ &: transition rule $\delta = H_i \xleftarrow{p_i:\alpha} B$ \\
        $\mathtt{S_1}$ &: all returned states must be subsumed by $\mathtt{S_1}$\\
        $\mathtt{S_2}$ &: goal state
    \end{tabular}
    }
    \Output{
    \begin{tabular}{ c l }
        $\mathtt{PreS}$ &: states that can reach $\mathtt{S_2}$ by following $\delta$
    \end{tabular}    
    }
    \nonl\hrulefill\\
	$\mathtt{PreS} := \emptyset$\;
	\ForEach{$\mathtt{S_2'}\subseteq\mathtt{S_2}$ and $\mathtt{H_i'}\subseteq H_i$ 
		s.t. $\theta = \unifier(\mathtt{S_2'}, \mathtt{H_i'})$ exists}{
		$\mathtt{S} := (\mathtt{S_2}\theta \backslash \mathtt{H_i'}\theta) \cup B\theta$ \tcp*{identify a pre-state $\mathtt{S}$}
		\ForEach{$l \in (\mathtt{S_2}\theta \backslash \mathtt{H_i'}\theta)$ and $l' \in H_i\theta \cup B\theta$ s.t. $unifier(l, l')$ exists }{
			add $l \neq l'$ to $\mathtt{S}$ \tcp*{add object inequalities to $\mathtt{S}$}
		}
		\textcolor{blue}{$\mathtt{S_{pre}} := mgs(\mathtt{S}, \mathtt{S_1})$} \tcp*{apply the $\mathtt{S_1}$ constraint}
		\textcolor{blue}{\ForEach{ $\mathtt{s_{pre}} \in \mathtt{S_{pre}}$}{
    		\lIf{$|terms(\mathtt{s_{pre}})| \leq$ b}{add $\mathtt{s_{pre}}$ to $\mathtt{PreS}$\tcp*{apply the state bound\footnotemark}} 
		}}
	}
% 	\Return $\mathtt{PreS}$ 
	\caption{\texttt{Regression}. This algorithm implements Equation~\ref{eq:pCTL REBEL update} \textcircled{1}. All returned states must be subsumed by $\mathtt{S_1}$, and can reach $\mathtt{S_2}$ by following the $\delta$ transition rule. All returned states must be legal and respect OI-subsumption. 
	}
	\label{algo:Regression}
\end{algorithm}
\footnotetext{The state bound $b\in\mathbb{N}$ is for infinite RMDPs. More details in Section~\ref{sec:theoretic results}. }

\subsection{$Q^p$-Rules Generation}
\label{sec:Q rules}

Given a state probability function $V^p$, this section computes $Q^p$ probabilities for state-action pairs (cf. Equation~\ref{eq:pCTL REBEL update} \textcircled{2}). 
$Q^p(s,a)$ denotes the probability that a formula is satisfied by a path that starts from $s$ and has the first action $a$. 
Similar to the state probability function $V^p$, the $Q^p$ function is an ordered set of $Q^p$-rules. 
These $Q^p$-rules are an intermediate representation, from which we can compute the next state probability function (cf. Section~\ref{sec:V rules}).

Deriving a $Q^p$ function is complex for two reasons. 
First, we must consider all transition rules that indicate different action postconditions. For example, action $\mathtt{move}/3$ has two possible outcomes, namely succeeding and failing. 
As could be expected, it is not sufficient to consider only the case where the action always succeeds as in Figure~\ref{fig:path}.
We must consider all transition rules in order to correctly calculate the $Q^p$-rules.
Second, an abstract action can \textit{diverge} and produce multiple preconditions. 
For example, Figure~\ref{fig:path} shows three different preconditions are derived by applying different substitutions to the $\delta_{\mathtt{move_1}}$ transition. It is required to consider all these preconditions. Fortunately, to resolve these issues, we could follow the procedure provided by the original REBEL, with limited adaptions.

\texttt{$Q^p$Rules} (Algorithm~\ref{algo:QRules}) implements the procedure of computing a set of $Q^p$-rules (cf. Equation~\ref{eq:pCTL REBEL update} \textcircled{2}). The process is as the follows.  
First, for each transition rule and current states in $V^p_t$, \texttt{$Q^p$Rules} computes a set of \textit{partial rules} (cf. line 2-5). This procedure results in multiple sets of partial rules, and each set considers one single postcondition of the action. 
Then, we must combine these partial rule sets to get a complete  $Q^p$-rule set. This is done by unifying the partial rules (cf. line 6-14). 
Finally, \texttt{$Q^p$Rules} returns a set of complete $Q^p$-rules.

 \texttt{$Q^p$Rules} modifies the \texttt{QRULES} algorithm in REBEL~\citep{Kersting:2004} in two ways, marked \textcolor{blue}{blue}. First, it provides support to the \textit{until} formulae $[\mathtt{S_1\;U\;S_2}]$ by passing an extra parameter $\mathtt{S_1}$ to Algorithm~\ref{algo:Regression} (cf. Section~\ref{sec:logical regression}). Second, it achieves the probabilistic interpretation of the Bellman operator by discarding the reward component and setting the discount factor to 1. 
\begin{algorithm}[ht]
    \SetKwInOut{Input}{Require}
    \SetKwInOut{Output}{Return}
    \Input{
    \begin{tabular}{ c l }
        $V^p_t$ &: state probability function \\
        $\mathtt{S_1}$ &: all states in the returned $Q^p$-rules must be subsumed by $\mathtt{S_1}$
    \end{tabular}
    }
    \Output{
    \begin{tabular}{ c l }
        ${Q^p}_{t+1}$ &: the next state-action probability function
    \end{tabular}    
    }
    \nonl\hrulefill\\
	${Q^p}_{t+1} := \emptyset$\;
	\tcc{fix an action $\alpha$}
	\ForEach{$H_i \xleftarrow{p_i:\alpha} B$ for $\alpha$}{
		\ForEach{$v^p\leftarrow V \in V^p_t$}{
		    $\mathtt{PreS}$ := \texttt{Regression}$(H_i \xleftarrow{p_i:\alpha} B,\textcolor{blue}{\mathtt{S_1}},V)$\;
		    \tcc{derive a subset of partial rules}
			$partialQ^p := \{q^p : \alpha\theta \leftarrow S | S \in \mathtt{PreS} \mbox{ and } \textcolor{blue}{q^p = p_i \times v^p} \}$\;
			\lIf{${Q^p}_{t+1} = \emptyset$}{
				${Q^p}_{t+1} := partialQ^p$ 
			}
			\uElse{
				$newQ^p := \emptyset$ \;
				\For{\mbox{ all pairs } $q^p_1:A_1\leftarrow S_1 \in {Q^p}_{t+1}$ and \\
				    \qquad\qquad\qquad$q^p_2:A_2\leftarrow S_2 \in partialQ^p$}
				{
				\tcc{unify the partial rules}
				\ForEach{$A\leftarrow S= mgs(A_1\leftarrow S_1, A_2\leftarrow S_2)$}{
					$\textcolor{blue}{q^p := q^p_1 + q^p_2}$ \;
					add $q^p:A\leftarrow S$ to $newQ^p$ 
				}
				${Q^p}_{t+1} := newQ^p$ 
				}
			}
		}
	}
	\caption{\texttt{$Q^p$Rules}. This algorithm implements Equation~\ref{eq:pCTL REBEL update} \textcircled{2}.
	All returned $Q^p$-rules must be legal and respect OI-subsumption. All states in the returned $Q^p$-rules must be subsumed by $\mathtt{S_1}$.
	}
	\label{algo:QRules}
\end{algorithm}

\subsection{$V^p$-Rules Generation}
\label{sec:V rules}

This section calculates the new state probability function $V^p_{t+1}$, given ${Q^p}_{t+1}$, by maximizing over the actions (cf. Equation~\ref{eq:pCTL REBEL update} \textcircled{3}). 
Recall that ${Q^p}_{t+1}$ is an ordered set of $Q^p$-rules of the form of $q^p:A\leftarrow S$. 
The task is to derive $V^p_{t+1}$, an ordered set of $V^p$-rules of the form of $v^p\leftarrow S$.

Trivially, turning ${Q^p}_{t+1}$ into $V^p_{t+1}$ takes three steps. First, the ${Q^p}$-rules must be sorted such that a rule connected to a high probability has a high priority as we are interested in the \textit{maximum} probability (as defined by Equation~\ref{eq: max probability}). 
Second, the redundant $Q^p$-rules must be removed. 
A $Q^p$-rule is redundant if it is subsumed by another $Q^p$-rule that has a higher priority.
Third, the remaining $Q^p$-rules are turned into $V^p$-rules by removing the action in the rule.

\texttt{$V^p$Rules} (Algorithm~\ref{algo:VRules}) implements Equation~\ref{eq:pCTL REBEL update} \textcircled{3}.
The process is as follows.
First, the $Q^p$-rules are ordered decreasingly so that a state is always assigned a \textit{maximum} probability (line 2). 
Second, 
to remove redundant $Q^p$-rules,
an \textit{absorbing rule} is required when the formula has an absorbing goal in that no more transitions occur after the goal is reached. Hence, given an absorbing goal, any rule concerning transitions that start from the goal is redundant and should be removed. 
Concretely,
an \textit{until} formula $[\mathtt{S_1\;U\; S_2}]$ has an absorbing goal $\mathtt{S_2}$ such that an execution stops once $\mathtt{S_2}$ is reached. Hence, an absorbing rule $1.0\leftarrow \mathtt{S_2}$ must be inserted to the beginning of the $Q^p$-rules. 
On the other hand, a \textit{next} formula $[\mathtt{X\; S_2}]$ does not have an absorbing goal as it is possible to follow exactly one transition from $\mathtt{S_2}$ to another state where $\mathtt{S_2}$ may or may not hold. 
Therefore, no absorbing rules are inserted.
Finally, redundant rules in the ordered set are removed (line 6-10).

\texttt{$V^p$Rules} generalizes the \texttt{VRULES} algorithm in REBEL~\citep{Kersting:2004} to handle non-absorbing \textit{next} formulae $[\mathtt{X\; S_2}]$ as the original REBEL considers only absorbing goals. The generalized part is marked \textcolor{blue}{blue}.

\begin{algorithm}[ht]
    \SetKwInOut{Input}{Require}
    \SetKwInOut{Output}{Return}
    \Input{
    \begin{tabular}{ c l }
        ${Q^p}_{t+1}$ &: state-action probability function \\
        $\psi$ &: pCTL formula $[\mathtt{S_1\;U\;S_2}]$ or $[\mathtt{X\; S_2}]$
    \end{tabular}
    }
    \Output{
    \begin{tabular}{ c l }
        $V^p_{t+1}$ &: the next state probability function
    \end{tabular}    
    }
    \nonl\hrulefill\\
	$V^p_{t+1} := \emptyset$ \; 
	sort ${Q^p}_{t+1}$ in decreasing order of $Q^p$ probability\;
	\tcc{Add absorbing $Q^p$-rules}
	\textcolor{blue}{
	\lIf{$\psi = \mathtt{S_1\;U^{\leq k}S_2}$}{${Q^p}_{abs} := \{1.0:\mathtt{\emptyset\leftarrow S | S\in S_2}\}$}
	\lElse{${Q^p}_{abs} := \emptyset$}
	}
	\textcolor{blue}{${Q^p}_{t+1} := {Q^p}_{abs} + {Q^p}_{t+1}$} \tcp*{attach absorbing $Q^p$-rules to the top of ${Q^p}_{t+1}$}
	\While{${Q^p}_{t+1} \neq \emptyset$}{
		remove the top $Q^p$-rule $d:A\leftarrow B$ from their ${Q^p}_{t+1}$ \; 
		\uIf{no other rule $d:A'\leftarrow B'$ in ${Q^p}_{t+1}$ exists s.t. $B'$ subsumes $B$}{
			add $d \leftarrow B$ to $V^p_{t+1}$ \;
			\tcc{remove redundant $Q^p$-rules}
			remove all rules $d'' \leftarrow B''$ from ${Q^p}_{t+1}$ s.t. $B$ subsumes $B''$ 
		}
	}
% 	\Return $V^p_{t+1}$
	\caption{\texttt{$V^p$Rules}. This function implements Equation~\ref{eq:pCTL REBEL update} \textcircled{3}. All returned $V^p$-rules must be legal and respect OI-subsumption.
	}
	\label{algo:VRules}
\end{algorithm}

\subsection{PCTL-REBEL Illustration}
\label{sec: illustration}

This section illustrates pCTL-REBEL (cf. Equation~\ref{eq:pCTL REBEL update}), namely, taking a state probability function $V^p_{t}$ to compute the next state probability function $V^p_{t+1}$. 
Clearly, different pCTL formulae require different numbers of iterations. That is, $[\mathtt{X\; S_2}]$ requires one iteration, $[\mathtt{S_1\;U^{\leq k}S_2}]$ requires $\mathtt{k}$ iterations, and $[\mathtt{S_1\;U\;S_2}]$ requires an arbitrary number iterations to obtain an accurate enough approximation. 
For simplicity, we illustrate with path formulae (without probabilities) that require one iteration. 
Section~\ref{sec:pCTL-REBEL} will cover the full pCTL language. 
\begin{description}
\item[\textbf{Formula 1} $\psi_1 = \mathtt{[X\;on(a,b)]}$:] Find all states that reach $\{\mathtt{on(a,b)}\}$ after 1 step.
\item[\textbf{Formula 2} $\psi_2 = \mathtt{[on(c,d)\;U^{\leq 1}on(a,b)]}$:] Find all states that reach $\{\mathtt{on(a,b)}\}$ within 1 step by going through only the states where $\{\mathtt{on(c,d)}\}$ holds.
\end{description}
We consider the $\delta_{\mathtt{move}}$ transition in the blocks world (also in Figure~\ref{fig:abstract transition}).
\begin{align*}
\delta_{\mathtt{move_1}}:
    \mathtt{cl(A){,}cl(C){,}on(A,B)
			\xleftarrow{0.9:move(A,B,C)}
			cl(A){,}cl(B){,}on(A,C)}\\
\delta_{\mathtt{move_2}}:
    \mathtt{cl(A){,}cl(B){,}on(A,C)
			\xleftarrow{0.1:move(A,B,C)}
			cl(A){,}cl(B){,}on(A,C)}
\end{align*}
% $$
% \delta_{\mathtt{move_1}} :
%     \mathtt{cl(A){,}cl(C){,}on(A,B)
% 			\xleftarrow{\makebox[1.8cm]{
% 			\begin{scriptsize}
% 			$\mathtt{0.9:move(A,B,C)}$
% 			\end{scriptsize}}}
% 			cl(A){,}cl(B){,}on(A,C)}
% $$
% $$
% \delta_{\mathtt{move_2}} :
%     \mathtt{cl(A){,}cl(B){,}on(A,C)
% 			\xleftarrow{\makebox[1.8cm]{
% 			\begin{scriptsize}
% 			$\mathtt{0.1:move(A,B,C)}$
% 			\end{scriptsize}}}
% 			cl(A){,}cl(B){,}on(A,C)}
% $$

\subsubsection*{\textbf{PCTL-REBEL on Formula 1: $[\mathtt{X\;on(a,b)}]$}}
For the formula $\psi_1 = [\mathtt{X\;on(a,b)}]$, the initial function $V^p_0$ is 
\begin{itemize}
\item[] $1.0 \leftarrow \mathtt{on(a,b)}$
\item[] $0.0 \leftarrow \emptyset$ 
\end{itemize}
\textcircled{1} 
\texttt{Regression}. Given the initial state probability function $V^p_0$, to obtain all possible pre-states, \texttt{Regression} is called with all combinations of transition rules and $V^p$-rules, e.g. \texttt{Regression}$(\delta_{\mathtt{move_1}}$, $\emptyset$, $\mathtt{on(a,b)})$ \footnote{The second argument is $\emptyset$ as the $\mathtt{X}$ operator is not absorbing.}. 
This results in two sets of partial $Q^p$-rules such that each set considers one outcome of the $\mathtt{move}$ action. 
The resulting partial $Q^p$-rules are listed below. The $\langle 1\cdot\rangle$ rules correspond to the successful outcome (i.e. $\delta_{\mathtt{move_1}}$) and the $\langle 2\cdot\rangle$ rules correspond to the unsuccessful outcome (i.e. $\delta_{\mathtt{move_2}}$).
Table~\ref{tab: combination} shows the the corresponding transitions and states. All partial $Q^p$-rules respect OI-subsumption.
\begin{align*}
\langle 1a\rangle &\;0.9:\mathtt{move(a,b,Z)\leftarrow cl(a),cl(b),on(a,Z)}\\
\langle 1b\rangle &\;0.9:\mathtt{move(X,a,Z)\leftarrow cl(X),cl(a),on(X,Z),on(a,b)}\\
\langle 1c\rangle &\;0.9:\mathtt{move(X,Y,a)\leftarrow cl(X),cl(Y),on(X,a),on(a,b)}\\
\langle 1d\rangle &\;0.9:\mathtt{move(X,Y,Z)\leftarrow cl(X),cl(Y),on(X,Z),on(a,b)}\\
\langle 1e\rangle &\;0.0:\mathtt{move(X,Y,Z)\leftarrow cl(X),cl(Y),on(X,Z)}\\
\langle 2a\rangle &\;0.1:\mathtt{move(a,Y,b)\leftarrow cl(a),cl(Y),on(a,b)}\\
\langle 2b\rangle &\;0.1:\mathtt{move(X,a,Z)\leftarrow cl(X),cl(a),on(X,Z),on(a,b)}\\
\langle 2c\rangle &\;0.1:\mathtt{move(X,Y,a)\leftarrow cl(X),cl(Y),on(X,a),on(a,b)}\\
\langle 2d\rangle &\;0.1:\mathtt{move(X,Y,Z)\leftarrow cl(X),cl(Y),on(X,Z),on(a,b)}\\
\langle 2e\rangle &\;0.0:\mathtt{move(X,Y,Z)\leftarrow cl(X),cl(Y),on(X,Z)}
\end{align*}
\begin{table}[h]
\begin{tabular}{c|c|c} 
    &   $\delta_{\mathtt{move_1}}$     & $\delta_{\mathtt{move_2}}$  \\\hline
$\mathtt{on(a,b)}$ 
& $\langle 1a\rangle \langle 1b\rangle \langle 1c\rangle \langle 1d\rangle$   
& $\langle 2a\rangle \langle 2b\rangle \langle 2c\rangle \langle 2d\rangle$ \\\hline
$\emptyset$ 
& $\langle 1e\rangle$   
& $\langle 2e\rangle$
\end{tabular}
\captionsetup{width=.7\linewidth}
\caption{Different combinations of transition rules and states result in different partial $Q^p$-rules $\langle 1a \rangle$ - $\langle 2e \rangle$. The states on the leftmost column comes from $V^p_0$. }
\label{tab: combination}
\end{table}

\noindent\textcircled{2} \texttt{$Q^p$Rules}. To compute the $Q^p$-rules, the two sets of partial $Q^p$-rules must be combined. To do so, we consider all possible combinations of $\langle 1\cdot \rangle$ and $\langle 2\cdot \rangle$. The resulting $Q^p$-rules are listed as follows, along with how they are created. All rules respect OI-subsumption.
\begin{align*}
    \langle 1\rangle &\;1.0:\mathtt{move(X,a,Z) \leftarrow cl(X), cl(a), on(X,Z), on(a,b)} & \langle 1b\rangle{+}\langle 2b\rangle\\
    \langle 2\rangle &\;1.0:\mathtt{move(X,Y,a) \leftarrow cl(X), cl(Y), on(X,a), on(a,b)} & \langle 1c\rangle{+}\langle 2c\rangle\\
    \langle 3\rangle &\;1.0:\mathtt{move(X,Y,Z) \leftarrow cl(X), cl(Y), on(X,Z), on(a,b)} & \langle 1d\rangle{+}\langle 2d\rangle\\
	\langle 4\rangle &\;0.9:\mathtt{move(a,b,Z) \leftarrow cl(a), cl(b), on(a,Z)} & \langle 1a\rangle{+}\langle 2e\rangle\\
    \langle 5\rangle &\;0.9:\mathtt{move(X,a,Z) \leftarrow cl(X), cl(a), on(X,Z), on(a,b)} & \langle 1b\rangle{+}\langle 2e\rangle\\
    \langle 6\rangle &\;0.9:\mathtt{move(X,Y,a) \leftarrow cl(X), cl(Y), on(X,a), on(a,b)} & \langle 1c\rangle{+}\langle 2e\rangle\\
	\langle 7\rangle &\;0.9:\mathtt{move(X,Y,Z) \leftarrow cl(X), cl(Y), on(X,Z), on(a,b)} & \langle 1d\rangle{+}\langle 2e\rangle\\
	\langle 8\rangle &\;0.1:\mathtt{move(a,Y,b) \leftarrow cl(a), cl(Y), on(a,b)} & \langle 1e\rangle{+}\langle 2a\rangle\\
	\langle 9\rangle &\;0.1:\mathtt{move(X,a,Z) \leftarrow cl(X), cl(a), on(X,Z), on(a,b)} & \langle 1e\rangle{+}\langle 2b\rangle\\
	\langle 10\rangle &\;0.1:\mathtt{move(X,Y,a) \leftarrow cl(X), cl(Y), on(X,a), on(a,b)} & \langle 1e\rangle{+}\langle 2c\rangle\\
	\langle 11\rangle &\;0.1:\mathtt{move(X,Y,Z) \leftarrow cl(X), cl(Y), on(X,Z), on(a,b)} & \langle 1e\rangle{+}\langle 2d\rangle\\
	\langle 12\rangle &\;0.0:\mathtt{move(X,Y,Z) \leftarrow cl(X), cl(Y), on(X,Z)} & \langle 1e\rangle{+}\langle 2e\rangle
\end{align*}

The $Q^p$-rules must be ordered by their probabilities as above. The rules $\langle 1\rangle$-$\langle 3\rangle$ are interchangeable as they have the same probability. Similarly, $\langle 4\rangle$-$\langle 7\rangle$ and $\langle 8\rangle$-$\langle 11\rangle$ are interchangeable, respectively.

\textcircled{3} \texttt{$V^p$Rules}. Now we can derive the new $V^p$-rules by removing redundant $Q^p$-rules and dropping the action components.
The resulting $V^p$-rules are shown below where the numbering inherits the one of the $Q^p$-rules. Rules $\langle 5\rangle$-$\langle 7\rangle$ are redundant because they are subsumed by $\langle 1\rangle$-$\langle 3\rangle$, respectively. Similarly, rules $\langle 9\rangle$-$\langle 11\rangle$ are redundant as they are subsumed by $\langle 1\rangle$-$\langle 3\rangle$, respectively.
\begin{align*}
    \langle 1\rangle &\;1.0\leftarrow \mathtt{cl(X), cl(a), on(X,Z), on(a,b)} \\
    \langle 2\rangle &\;1.0\leftarrow \mathtt{cl(X), cl(Y), on(X,a), on(a,b)} \\
    \langle 3\rangle &\;1.0\leftarrow \mathtt{cl(X), cl(Y), on(X,Z), on(a,b)} \\
    \langle 4\rangle &\;0.9\leftarrow \mathtt{cl(a), cl(b), on(a,Z)}\\
    \langle 8\rangle &\;0.1\leftarrow \mathtt{cl(a), cl(Y), on(a,b)} \\
    \langle 12\rangle &\;0.0\leftarrow \mathtt{cl(X), cl(Y), on(X,Z)}
\end{align*}
Given $\psi_1=[\mathtt{X\; on(a,b)}]$, and the initial state probability function $V^p_0$, we have applied pCTL-REBEL and obtained $V^p_1$, which assigns to all states a maximum probability of reaching $\mathtt{on(a,b)}$ after exactly one step.

\subsubsection*{\textbf{PCTL-REBEL on Formula 2: $[\mathtt{on(c,d)\;U^{\leq 1}on(a,b)}]$}}
For the formula $\psi_2 = [\mathtt{on(c,d)\;U^{\leq 1}on(a,b)}]$, the initial probability function $V^p_0$ is 
\begin{itemize}
\item[] $1.0 \leftarrow \mathtt{on(a,b)}$
\item[] $0.0 \leftarrow \emptyset$ 
\end{itemize}
\textcircled{1} 
\texttt{Regression}. Given the initial state probability function $V^p_0$, to obtain all possible pre-states, \texttt{Regression} is called with all combinations of transition rules and $V^p$-rules, e.g. \texttt{Regression}$(\delta_{\mathtt{move_1}}$, $\mathtt{on(c,d)}$, $\mathtt{on(a,b)})$. 
This results in two sets of partial $Q^p$-rules such that each set considers one outcome of the $\mathtt{move}$ action.
We list some of the resulting partial $Q^p$-rules below. 
The $\langle 1\cdot\rangle$ rules correspond to the successful outcome (i.e. $\delta_{\mathtt{move_1}}$) and the $\langle 2\cdot\rangle$ rules correspond to the unsuccessful outcome (i.e. $\delta_{\mathtt{move_2}}$).
Table~\ref{tab: combination2} shows the the corresponding transitions and states. All partial $Q^p$-rules respect OI-subsumption.
\begin{align*}
\langle 1a\rangle &\;0.9:\mathtt{move(a,b,c)\leftarrow cl(a),cl(b),on(a,c),on(c,d)}\\
\langle 1b\rangle &\;0.9:\mathtt{move(a,b,Z)\leftarrow cl(a),cl(b),on(a,Z),on(c,d)}\\
\langle 1c\rangle &\;0.9:\mathtt{move(c,a,d)\leftarrow cl(a),cl(c),on(a,b),on(c,d)}\\
\langle 1d\rangle &\;0.9:\mathtt{move(c,Y,d)\leftarrow cl(c),cl(Y),on(a,b),on(c,d)}\\
\cdots\\
\langle 1l\rangle &\;0.0:\mathtt{move(X,c,Z)\leftarrow cl(X),cl(c),on(X,Z),on(c,d)}\\
\langle 1m\rangle &\;0.0:\mathtt{move(c,Y,d)\leftarrow cl(c),cl(Y),on(c,d)}\\
\langle 1n\rangle &\;0.0:\mathtt{move(X,Y,c)\leftarrow cl(X),cl(Y),on(X,c),on(c,d)}\\
\langle 1o\rangle &\;0.0:\mathtt{move(X,Y,Z)\leftarrow cl(X),cl(Y),on(X,Z),on(c,d)}\\
\langle 2a\rangle &\;0.1:\mathtt{move(c,a,d)\leftarrow cl(a),cl(c),on(a,b),on(c,d)}\\
\langle 2b\rangle &\;0.1:\mathtt{move(c,Y,d)\leftarrow cl(c),cl(Y),on(a,b),on(c,d)}\\
\cdots\\
\langle 2m\rangle &\;0.0:\mathtt{move(c,Y,d)\leftarrow cl(c),cl(Y),on(c,d)}\\
\langle 2n\rangle &\;0.0:\mathtt{move(X,Y,c)\leftarrow cl(X),cl(Y),on(X,c),on(c,d)}\\
\langle 2o\rangle &\;0.0:\mathtt{move(X,Y,Z)\leftarrow cl(X),cl(Y),on(X,Z),on(c,d)}\\
\end{align*}
\begin{table}[h]
\begin{tabular}{c|c|c} 
    &   $\delta_{\mathtt{move_1}}$     & $\delta_{\mathtt{move_2}}$  \\\hline
$\mathtt{on(a,b)}$ 
& $\langle 1a\rangle$-$\langle 1k\rangle$   
& $\langle 2a\rangle$-$\langle 2k\rangle$ \\\hline
$\emptyset$ 
& $\langle 1l\rangle$-$\langle 1o\rangle$   
& $\langle 2l\rangle$-$\langle 2o\rangle$
\end{tabular}
\captionsetup{width=.7\linewidth}
\caption{Different combinations of transition rules and states result in different partial $Q^p$-rules $\langle 1a \rangle$ - $\langle 2o \rangle$. The states on the leftmost column comes from $V^p_0$. }
\label{tab: combination2}
\end{table}

\noindent\textcircled{2} 
\texttt{$Q^p$Rules}. To compute the $Q^p$-rules, the two sets of partial $Q^p$-rules must be combined. To do so, we consider all possible combinations of $\langle 1\cdot \rangle$ and $\langle 2\cdot \rangle$. Some of the resulting $Q^p$-rules are listed below, along with how they are created. All rules respect OI-subsumption.
\begin{align*}
    \langle 1\rangle &\;1.0:\mathtt{move(c,a,d) \leftarrow cl(a), cl(c), on(a,b),on(c,d)} & \langle 1c\rangle{+}\langle 2a\rangle\\
	\langle 2\rangle &\;1.0:\mathtt{move(c,Y,d) \leftarrow cl(c), cl(Y), on(a,b),on(c,d)} & \langle 1d\rangle{+}\langle 2b\rangle\\
	\cdots\\
	\langle  10\rangle &\;0.9:\mathtt{move(a,b,c) \leftarrow cl(a), cl(b), on(a,c),on(c,d)} & \langle 1a\rangle{+}\langle 2n\rangle\\
    \langle  11\rangle &\;0.9:\mathtt{move(a,b,Z) \leftarrow cl(a), cl(b), on(a,Z),on(c,d)} & \langle 1b\rangle{+}\langle 2o\rangle\\
    \cdots\\
    \langle 33\rangle &\;0.0:\mathtt{move(c,Y,d) \leftarrow cl(c), cl(Y), on(c,d)} & \langle 1m\rangle{+}\langle 2m\rangle\\
    \langle 34\rangle &\;0.0:\mathtt{move(X,Y,c) \leftarrow cl(X), cl(Y), on(X,c),on(c,d)} & \langle 1n\rangle{+}\langle 2n\rangle\\
	\langle 35\rangle &\;0.0:\mathtt{move(X,Y,Z) \leftarrow cl(X), cl(Y), on(X,Z),on(c,d)} & \langle 1o\rangle{+}\langle 2o\rangle
\end{align*}
The $Q^p$-rules must be ordered by their probabilities as above.

\textcircled{3} \texttt{$V^p$Rules}. Now we can derive the new $V^p$-rules.
Since $[\mathtt{on(a,b)}]$ is absorbing, the following absorbing rule must be inserted to the beginning of the $Q^p$-rule set. 
\begin{align*}
    \langle 0\rangle &\;1.0:\mathtt{\emptyset \leftarrow on(a,b)}
\end{align*}
After removing redundant $Q^p$-rules and dropping the action components in the rules, we obtain the resulting $V^p$-rules below. The numbering of the $V^p$-rules inherits the one of the $Q^p$-rules.
Most redundant rules are subsumed by the absorbing rule $\langle 0\rangle$.
\begin{align*}
	 \langle 0\rangle&\; 1.0\leftarrow\mathtt{on(a,b)}\\
	 \langle 10\rangle &\;0.9\leftarrow\mathtt{ cl(a), cl(b), on(a,c),on(c,d)} \\
    \langle 11\rangle &\;0.9\leftarrow\mathtt{ cl(a), cl(b), on(a,Z),on(c,d)} \\
    \langle 33\rangle &\;0.0\leftarrow\mathtt{ cl(c), cl(Y), on(c,d)} \\
    \langle 34\rangle &\;0.0\leftarrow\mathtt{ cl(X), cl(Y), on(X,c),on(c,d)}\\
	\langle 35\rangle &\;0.0\leftarrow\mathtt{ cl(X), cl(Y), on(X,Z),on(c,d)} 
\end{align*}

\section{Main Contribution -- a Relational Model Checker}
\label{sec:pCTL-REBEL}

This section introduces the main algorithm in this paper, the relational model checking algorithm, that solves the relational model checking problem (Section~\ref{sec:problem statement}). Formally, given a pCTL formula $\phi$ and an RMDP $K$, the relational model checker identifies the set $Sat_K(\phi)$ of abstract states that represent all ground states that satisfy $\phi$. 
Since the relational model checker is based on PCTL-REBEL (Section~\ref{sec:relational value iteration}), it operates at the relational level.

Different from Section~\ref{sec:relational value iteration} that computes one single iteration, this section allows for full relational pCTL formulae that requires multiple iterations and can be \textit{nested}. It is standard practice to represent a nested pCTL formula as a parse tree~\citep{Baier:2008}. In a parse tree, each leaf node is an abstract state and each inner node contains exactly one operator. 

\begin{example}
\label{ex: nested formula}
A nested relational pCTL formula example is as follows. 
\begin{align*}
\phi_{\mathtt{nested}}=& \mathtt{P_{\geq 0.5}[cl(a)\;U^{\leq 4}
(on(a,b)}
\mathtt{\wedge P_{\geq 0.8} [P_{\geq 0.9}[X\;cl(e)]\;U^{\leq 2}on(c,d) ])]}
\end{align*}
The nested formula $\phi_{\mathtt{nested}}$ is complex at first sight, however, it can be represented as a \textit{parse tree} in Figure~\ref{fig:parse tree}.
A state $s$ satisfies $\phi_{\mathtt{nested}}$ if and only if $s$ satisfies all the following three conditions. (1) A path starting from $s$ must reach $\{\mathtt{on(a,b)}\}$ within 4 steps with a probability greater than or equal to 0.5 by going through the states where $\{\mathtt{cl(a)}\}$ holds. (2) Then, the path must reach $\{\mathtt{on(c,d)}\}$ within 2 steps with a probability greater than or equal to 0.8 by going through the states that satisfy $\mathtt{P_{\geq 0.9}[X\;cl(e)]}$. (3) The states that satisfy $\mathtt{P_{\geq 0.9}[X\;cl(e)]}$ are the ones that can transition to $\{\mathtt{cl(e)}\}$ with a probability greater than or equal to 0.9 after exactly one step. 

\begin{figure}
\centering
\includegraphics[width=.7\textwidth]{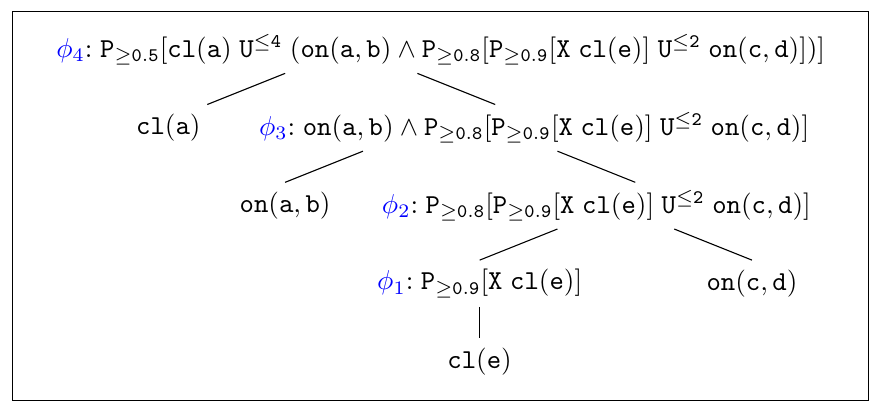}
\captionsetup{width=.7\linewidth}
\caption{The parse tree of $\phi_{\mathtt{nested}}$. Each inner node of the parse tree is annotated with a subformula $\phi_i$ and handles one operator. $\phi_1$ handles $\mathtt{X}$, $\phi_2$ handles $\mathtt{U^{\leq 2}}$, $\phi_3$ handles $\land$ and $\phi_4$ handles $\mathtt{U^{\leq 4}}$.} \label{fig:parse tree}
\end{figure}
\end{example}

PCTL-REBEL always use a parse tree to evaluate a given formula. A parse tree is recursively evaluated \textit{upwards}. That is, an inner node considers its child leaf nodes to evaluate a subformula $\phi_{i}$, resulting in a set of states  $Sat(\phi_{i})$. Then, the inner node \textit{folds} the sub-tree so that its parent node can be activated. The final set of states $Sat(\phi)$ is produced by the root. For example, in Figure~\ref{fig:parse tree}, an intermediate state set $Sat(\phi_{i})$ is used to evaluate $Sat(\phi_{i+1})$. The state set $Sat(\phi_{4})$ is returned as the solution.

Section~\ref{sec: pCTL-REBEL algo} defines the pCTL-REBEL model checking algorithm, which is composed of 3 mutually recursive algorithms. Then, Section~\ref{sec: properties of pCTL-REBEL} gives an overview of the properties of the model checking algorithm.

\subsection{The pCTL-REBEL Model Checking Algorithm}
\label{sec: pCTL-REBEL algo}

The pCTL-REBEL model checking algorithm reforms the satisfiability relation $\models^\theta$ (cf. Section~\ref{sec:pCTL}). 
Given an RMDP $K$ and a pCTL formula $\phi$, the pCTL-REBEL model checker $Sat_K(\phi)$, i.e. all states in $K$ that satisfy $\phi$.
The model checker consists of 3 algorithms \texttt{Check}, \texttt{CheckUntil} and \texttt{CheckNext} that call one another recursively. In particular, \texttt{Check} is the main algorithm and is mutually recursive with the other two algorithms.
\texttt{CheckUntil} handles the \textit{until} formulae of the form of $\mathtt{P_{\bowtie p}[\phi_1\;U^{\leq k}\phi_2]}$. \texttt{CheckNext} handles the \textit{next} formulae of the form of $\mathtt{P_{\bowtie p}[X\;\phi_2]}$.

Given a formula $\phi$, \texttt{Check} (Algorithm~\ref{algo: Check}) evaluates $\phi$ in a recursive fashion as follows.
If $\phi$ is a relation atom $l$ (resp. $\neg l$), it returns a single abstract state $\{l\}$ (resp. $\{\neg l$\}) (line 1-2). 
If $\phi$ is a conjunction of the form of $\phi_1 \land \phi_2$, it first computes $\phi_1$ and $\phi_2$ separately to get two sets of abstract states $Sat1$ and $Sat2$ (line 4-5). 
Then, for each possible pair of abstract states $s_1\in Sat1$ and $s_2\in Sat2$, it collects all maximally general specializations $mgs(s_1, s_2)$ (line 6). Since all $mgs(s_1, s_2)$ are OI-subsumed by $s_1$ and $s_2$ by definition, they automatically satisfy $\phi_1 \land \phi_2$. 
If $\phi$ is a disjunction of the form of $\phi_1 \lor \phi_2$, it returns the union of the solutions to $\phi_1$ and $\phi_2$ (line 7-8).
% When the given formula is of the form $\neg\phi_1$, it performs model checking on $\phi_1$ to get $Sat1$ containing all states that satisfy $\phi_1$. Then, it 
If $\phi$ is an \textit{until} formula, it calls \texttt{CheckUntil} (Algorithm~\ref{algo: check until}) (line 9-10).
If $\phi$ is a \textit{next} formula,, it calls \texttt{CheckNext} (Algorithm~\ref{algo: check next}) (line 11-12). 
% After value iteration, the two algorithm eliminate the states that do not satisfy $\phi$ within the given probability bounds.

\begin{algorithm}[ht]
    \SetKwInOut{Input}{Require}
    \SetKwInOut{Output}{Return}
    \Input{
    \begin{tabular}{ c l }
        $\phi$ &: pCTL formula
    \end{tabular}
    }
    \Output{
    \begin{tabular}{ c l }
        $Sat_K(\phi)$ &: abstract states that satisfy $\phi$
    \end{tabular}    
    }
    \nonl\hrulefill\\
	\lIf{$\phi$ = l}{$Sat_K(\phi)$ =  \{l\}}
	\lElseIf{$\phi$ = $\neg l$}{$Sat_K(\phi)$ =  \{$\neg l$\}}
	\uElseIf{$\phi$ = $\phi_1 \land \phi_2$}{
	    $Sat_1 = \mathtt{Check(\phi_1)}$\\
	    $Sat_2 = \mathtt{Check(\phi_2)}$\\
		$Sat_K(\phi)$ = $\{mgs(s1, s2)\;|\; s1\in Sat_1, s2\in Sat_2\}$ \;
	}
	\uElseIf{$\phi$ = $\phi_1 \lor \phi_2$}{
		$Sat_K(\phi) = \mathtt{Check(\phi_1)}\cup \mathtt{Check(\phi_2)}$ \;
	}
	\uElseIf{$\phi$ = $\mathtt{P}_{\bowtie p}[\phi_1 \mathtt{U^{\leq k}}\phi_2]$}{
		$Sat_K(\phi)$ = \texttt{CheckUntil}($\phi$)
	}
	\uElseIf{$\phi$ = $\mathtt{P}_{\bowtie p}[\mathtt{X}\;\phi_2]$}{
		$Sat_K(\phi)$ =  \texttt{CheckNext}($\phi$)
	}
	\caption{\texttt{Check}. The main algorithm to solve the relational model checking problem. This algorithm implements the satisfiability relation (Section~\ref{sec:pCTL}). \texttt{Check} is mutually recursive with \texttt{CheckUntil} and \texttt{CheckNext}. }
	\label{algo: Check}
\end{algorithm}

\texttt{CheckUntil} (Algorithm~\ref{algo: check until}) is mutually recursive with \texttt{Check} (Algorithm~\ref{algo: Check}). It computes a formula of the form of $\mathtt{P_{\bowtie p}[ \phi_1U^{\leq k}\phi_2]}$ or $\mathtt{P_{\bowtie p}[ \phi_1 U\phi_2]}$ where $\mathtt{\phi_1}$ and $\mathtt{\phi_2}$ are pCTL formulae.
To begin the process, \texttt{CheckUntil} computes $\mathtt{\phi_1}$ and $\mathtt{\phi_2}$ separately to get two sets of abstracts states (line 1-2). Then, it sets the step bound for pCTL-REBEL (line 3).
When the given formula has a bounded $\mathtt{U^{\leq k}}$ operator, the step bound is $\mathtt{k}$ . Otherwise, the step bound is set to infinity and an arbitrary number of pCTL-REBEL iterations are applied until convergence (line 5-11). 
The convergence condition is twofold. First, the abstract states in the probability function do not change, i.e. states in $V^p_t$ and $V^p_{t-1}$ are the same. Second, the state probabilities have converged with respect to a given threshold $\epsilon$, i.e. $\max_{s\in V^p_t}|V^p_{t}(s)-V^p_{t-1}(s)|<\epsilon$. 
Then, \texttt{CheckUntil} collects and returns the abstract states in $V^p_t$ that satisfy the given probability threshold $\mathtt{\bowtie p}$ (line 12-14). 
	
\begin{algorithm}[ht]
    \SetKwInOut{Input}{Require}
    \SetKwInOut{Output}{Return}
    \Input{
    \begin{tabular}{ c l }
        $\phi$ &: pCTL formula $\mathtt{P_{\bowtie p}[ \phi_1U^{\leq k}\phi_2]}$ or $\mathtt{P_{\bowtie p}[ \phi_1 U\phi_2]}$
    \end{tabular}
    }
    \Output{
    \begin{tabular}{ c l }
        $Sat_K(\phi)$ &: abstract states that satisfy $\phi$
    \end{tabular}    
    }
    \nonl\hrulefill\\
	$\mathtt{S_1} :=$ \texttt{Check}$(\phi_1)$\; 
	$\mathtt{S_2} :=$ \texttt{Check}$(\phi_2)$\;
	\lIf{$\mathtt{k}$ is not given}{$\mathtt{k}:=\infty$ \tcp*{set k as infinity for unbounded U} }
	$V^p_0 := \{1.0\gets s|s\in \mathtt{S_2}\} \cup \{0\gets \emptyset\}$ \tcp*{Initialize $V^p_0$} 
	$t := 0$ \tcp*{t is the number of iterations so far}
	$\epsilon' := \infty$ \tcp*{$\epsilon'$ is the distance between $V^p_t$ and $V^p_{t+1}$}
	\While{$t < \mathtt{k}$ or $\epsilon' \geq \epsilon$}{
    	    $V^p_{t+1}$ := \texttt{OneIteration}$(V^p_t, \mathtt{S_1\;U^{\leq k}S_2})$\;
    	    $\epsilon' := V^p_{t+1}$ - $V^p_{t}$\; 
    	    $t := t + 1$
	    }
	$Sat_K(\phi) := \emptyset$\;
	\ForEach{$p_i \gets s_i \in V^p_t$}{
		\lIf{$p_i \bowtie \mathtt{p}$}{
			add $s_i$ to $Sat_K(\phi)$
		}
	}
% 	\Return $Sat_K(\phi)$
	\caption{\texttt{CheckUntil}. 
	\texttt{CheckUntil} is mutually recursive with \texttt{Check}. }
	\label{algo: check until}
\end{algorithm}

\texttt{CheckNext} (Algorithm~\ref{algo: check next}) is mutually recursive with \texttt{Check} (Algorithm~\ref{algo: Check}). It computes a formula of the form of $\phi = \mathtt{P_{\bowtie p}[ X\;\phi_2]}$ where $\phi_2$ is another pCTL formula. 
\texttt{CheckNext} is a similar but simpler than \texttt{CheckUntil}.
It performs one single value iteration to obtain the probability function $V^p_1$ (line 2-3). Then it collects and returns the abstract states in $V^p_1$ that satisfy the given probability threshold $\mathtt{\bowtie p}$ (line 4-6). 
\texttt{CheckNext} needs only one iteration as it considers exactly one transition.

\begin{algorithm}[ht]
    \SetKwInOut{Input}{Require}
    \SetKwInOut{Output}{Return}
    \Input{
    \begin{tabular}{ c l }
        $\phi$ &: pCTL formula $\mathtt{P_{\bowtie p}[\mathtt{X}\; \phi_2]}$
    \end{tabular}
    }
    \Output{
    \begin{tabular}{ c l }
        $Sat_K(\phi)$ &: abstract states that satisfy $\phi$
    \end{tabular}    
    }
    \nonl\hrulefill\\
	$\mathtt{S_2}$ = \texttt{Check}$(\phi_2)$\;
	$V^p_0 := \{1.0\gets s|s\in \mathtt{S_2}\} \cup \{0\gets \emptyset\}$\;
	$V^p_1$ := \texttt{OneIteration}$(V^p_0,  \mathtt{X\; S_2})$\;
	$Sat_K(\phi) := \emptyset$\;
	\ForEach{$p_i \gets s_i \in V^p_1$}{
		\lIf{$p_i \bowtie \mathtt{p}$}{
			add $s_i$ to $Sat_K(\phi)$
		}
	}
% 	\Return $Sat_K(\phi)$
	\caption{
	\texttt{CheckNext}.
	\texttt{CheckNext} is mutually recursive with \texttt{Check}.}
	\label{algo: check next}
\end{algorithm}

\subsection{Properties of PCTL-REBEL}
\label{sec: properties of pCTL-REBEL}

PCTL-REBEL is a relational model checking algorithm that finds all states that satisfy a given pCTL formula. We discuss the properties of pCTL-REBEL. 

\begin{description}
\item[\textbf{Lifted}] 
instead of operating at the ground level, pCTL-REBEL performs lifted inference as both the formula and the states are specified at an abstract level using relational representations.
Using lifted inference allows pCTL-REBEL to exploit relational symmetries in the model and make abstraction of the domain, hence mitigate the state explosion problem. The lifted inference was discussed in detail in Section~\ref{sec:relational value iteration}. 

\item[\textbf{Sound for step-bounded pCTL formulae}]
PCTL-REBEL is sound for finite RMDPs (that have a finite domain) and any step-bounded pCTL formulae. 
PCTL-REBEL is not sound for indefinite-horizon formulae (e.g. $\mathtt{U}$), just like many other value iteration algorithms. This is because pCTL-REBEL uses a naive termination criterion with some arbitrary convergence threshold $\epsilon$ (see \texttt{CheckUntil}, Algorithm~\ref{algo: check until}). Although this naive termination criterion is not sound, it achieves precise approximation in practice. For further details, please refer to~\cite{Haddad2014}.

\item[\textbf{Complete}]
PCTL-REBEL is complete for finite RMDPs such that the state probability function $V^p$ assigns a probability to all states. PCTL-REBEL captures the entire state space by using an ordered set of relational $V^p$-rules. Those states that are not captured by any other $V^p$-rules are guaranteed to be covered by the last $V^p$-rule $0\leftarrow\emptyset$ (see Section~\ref{sec:relational Bellman operator}).
\end{description}

\section{Relational Model Checking for Infinite MDPs}
\label{sec:theoretic results}

It is clear that when the domain is finite, the model checking problem is decidable as one can enumerates all states in the model, as discussed in Section~\ref{sec:problem statement}.
However, when the domain is infinite, the state space is typically infinite, making enumerating all states infeasible. 
In this section, we obtain decidability for a special class of infinite RMDPs. 
We will prove that under the \textit{state-boundedness} condition~\citep{Belardinelli2011}, a \textit{finite abstraction} of such RMDPs can be constructed checked by pCTL-REBEL. 
The main idea is to generate a finite abstraction that captures all relevant information of the underlying infinite RMDP with respect to a pCTL formula. Accordingly, checking the abstraction is equivalent to checking the infinite RMDP. 
By checking the finite abstraction, the model checking problem becomes decidable.
This section adopts the approach of~\citet{Belardinelli2011} to construct finite abstractions of infinite RMDPs. 
Furthermore, we show that pCTL-REBEL can naturally handle such abstractions as they are structurally similar to RMDPs.

Section~\ref{sec: abstract MDPs} defines the state-boundedness condition and under which, the finite abstraction of an infinite RMDP. Section~\ref{sec: probabilistic Bisimulation} proves that checking a pCTL formula $\phi$ against the finite abstraction is equivalent to checking $\phi$ against the corresponding infinite RMDP. Section~\ref{sec: pCTL REBEL infinite MDPs} discusses properties of pCTL-REBEL when handling such infinite RMDPs.

\subsection{Abstract MDP: a Finite Abstraction of an Infinite RMDP}
\label{sec: abstract MDPs}

Given an infinite RMDP, the relational model checking problem is generally undecidable due to the possibly infinite domain.
To obtain decidability, this section constructs a finite abstraction of a given infinite RMDP, called an \textit{abstract RMDP} (ARMDP). 
The purpose of an ARMDP is to use a finite model to capture all relevant information about a pCTL formula.
An ARMDP must be constructed under the \textit{state-boundedness} condition. 

The state-boundedness condition states that any state concerns only a finite number of objects.
For example, consider a blocks world that has infinitely many blocks and a table with a capacity $b$. An agent can take a block away or put a new block to the table, but no more than $b$ blocks can be on the table at any moment. Hence, the state bound is $b$. Since any two states can describe totally different blocks, the model still contains \textit{infinitely many} states. We say the blocks on the table are in the \textit{active domain}. We now formally define active domain and state-boundedness. 

\begin{define}
A bounded state $s_b$ is a finite subset of a ground state $s$ in some RMDP, i.e. $s_b\subseteq s$. The active domain of $s_b$, denoted $\adom(s_b)$, is the set of all domain objects in $s_b$. 
\end{define}
An active domain $\adom(s_b)$ is by definition finite as a bounded state $s_b$ is finite. A bounded state is similar to a ground state but concerns only a finite number of objects. That is, all atoms in $s_b$ are $\true$ and all the others are $\false$.

\begin{define}
Given an RMDP $K=\langle \Sigma,\Delta\rangle$, its underlying MDP $M = \langle S, A, T\rangle$, a state bound $b$, and a starting state $s_0$ such that $|\adom(s_0)| \leq b$, 
an MDP $M_b = \langle S_b, A_b, T_b\rangle$ can be defined by including all states that are reachable from $s_0$ and contain at most $b$ constants. 
Formally, $M_b = \langle S_b, A_b, T_b\rangle$ is defined as 
\begin{align*}
    S_b &:= \{s \;|\; s\in S, |\adom(s)| \leq b\}\\
    T_b &:= \{h\xleftarrow{p:a} b\in T| h\in S_b, b\in S_b\}\\
    A_b &:= \{a | h\xleftarrow{p:a} b\in T_b\}\\
\end{align*}
If $b\in\mathbb{N}$, then $M_b$ is called state-bounded or $b$-bounded.
\end{define}
A b-bounded MDP $M_b$ is uniquely defined by an RMDP and a state bound.
Roughly speaking, $M_b$ a \textit{sub-MDP} of $M$ that concerns at most $b$ objects in any state. 

\begin{example}
\label{ex: state-bounded MDP}
Consider a blocks world $K=\langle \Sigma, \Delta\rangle$ with $\Sigma = \langle R, D\rangle$ where 
$R=\mathtt{\{cl/1, on/2\}}$, $D=\{\mathtt{bl_i}|i\in \mathbb{N}\}$ is infinite, and $\Delta$ contains the following rules (as in Figure~\ref{fig:abstract transition}). 
\begin{align*}
\delta_{\mathtt{move_1}}:
    \mathtt{cl(A){,}cl(C){,}on(A,B)
			\xleftarrow{0.9:move(A,B,C)}
			cl(A){,}cl(B){,}on(A,C)}\\
\delta_{\mathtt{move_2}}:
    \mathtt{cl(A){,}cl(B){,}on(A,C)
			\xleftarrow{0.1:move(A,B,C)}
			cl(A){,}cl(B){,}on(A,C)}
\end{align*}
Given a state bound $b$, the RMDP $K$ defines a $b$-bounded MDP $M_b = \langle S_b, A_b, T_b\rangle$ such that each bounded state $s_b\in S_b$ contains at most $b$ blocks. 
We will use this blocks world example throughout this section.
\end{example}

Having defined state-boundedness, let us now move on to constructing an abstract MDP (ARMDP).
With respect to a pCTL formula, some domain objects in an MDP are irrelevant. For example, $\mathtt{P_{\geq 0.9}[X\;cl(a)]}$ does not concern any particular blocks other than block $\mathtt{a}$. 
By abstracting away irrelevant objects of a pCTL formula $\phi$, we can construct an ARMDP $M_\phi$ that captures all necessary information to check $\phi$. 
% Furthermore, checking $\phi$ against $M_\phi$ \textit{is equivalent to} checking $\phi$ against $M_b$.
Concretely, to construct an ARMDP, all \textit{$\phi$-relevant} domain objects are preserved, and all other objects are abstracted using variables. 
A domain object is \textit{$\phi$-relevant} if and only if it is in $\phi$ or in a transition rule. 
Furthermore, if a formula $\phi$ concerns a finite number of objects, the corresponding ARMDP $M_\phi$ is finite. 
In this paper, we assume all formulae contain at most $b$ objects where $b$ is the state bound. 

\begin{define} 
\label{def: ARMDP}
For a $b$-bounded MDP $M_b = \langle S_b, A_b, T_b \rangle$ of an RMDP $K=\langle\Sigma,\Delta\rangle$ and a relational pCTL formula $\phi$,
a $b$-bounded abstract RMDP $M_\phi = \langle S_\phi, \Sigma, \Delta, W\rangle$ is defined where
\begin{align*}
    S_\phi = \{s_\phi | 
    &\exists s\in S_b.\; s \preceq_{\theta} s_\phi, \\
    &consts(s_\phi)\subseteq consts(\phi)\cup consts(\Delta), \\
    &vars(s_\phi)\subseteq W
    \} 
\end{align*}
and $W$ is a set of $b-|consts(\phi)\cup consts(\Delta)|$ distinct variables. 
All terms in $s_\phi\in S_\phi$ are from a finite set of terms 
$$
terms(s_\phi) := W \cup consts(\phi)\cup consts(\Delta)
$$ 
It is assumed that $vars(\phi)\in W$. It is assumed that states in $S_\phi$ are not syntactic variants, that is, they are not a variable renaming of one another. 
\end{define}

An ARMDP is finite as only a finite number of terms are allowed in the state description.
An ARMDP is $b$-bounded as each state $s_\phi\in S_\phi$ contains at most $b$ terms, i.e. $|terms(s_\phi)| \leq b$. 
Since an ARMDP state is $b$-bounded, it has bounded branching behavior such that the number of available actions in any state is finite. 
The finite state space of ARMDPs is crucial to obtaining decidability. 

An ARMDP has a similar structure as an RMDP but has a more abstract state space. That is, unlike RMDPs, the state space of an ARMDP is not connected to an explicit domain as any ARMDP state concerns only $\phi$-relevant constants and variables. By replacing the variables by domain constants, an ARMDP state can enumerate infinitely many ground states in the underlying RMDP.

\begin{example}
\label{ex: ARMDP} (Cont. Example~\ref{ex: state-bounded MDP})
A $b$-bounded MDP $M_b=\langle S_b, A_b, T_b\rangle$ and the formula $\phi=\mathtt{cl(a)}$ defines an ARMDP $M_\phi = \langle S_\phi, \Sigma, \Delta, W\rangle$. A state $s_b\in S_b$ concerns at most $b$ terms, namely $\{ \mathtt{Bl_1, Bl_2, ..., Bl_{b-1}, a}\}$.
The abstract states $s_1=\{\mathtt{cl(a)}\}\in S_\phi$ and $s_2=\{\mathtt{cl(Bl_1)}\}\in S_\phi$ represent the set of all infinitely many ground states that have at least one clear block. These ground states can be enumerated by assigning domain objects to $\mathtt{Bl_1}$.
\end{example}

This section has shown that given a state-bounded MDP and a pCTL formula, an finite abstraction can be constructed. Such abstraction is called an ARMDP. An ARMDP is similar to an RMDP but is state-bounded, finite, and captures only $\phi$-relevant information. 
The next section will show that checking an ARMDP is equivalent to checking the underlying model.

\subsection{Decidable Model Checking for ARMDPs}
\label{sec: probabilistic Bisimulation}

This section proves the decidability of the model checking problem for a special class of infinite RMDPs, namely, the ARMDP, defined in Section~\ref{sec: abstract MDPs}. Decidability is obtained by proving that checking the ARMDP yields equivalent results as checking the underlying infinite, state-bounded MDP, namely, Theorem~\ref{theorem1}.

\begin{theorem} 
\label{theorem1}
    For a $b$-bounded MDP $M_b$ and its corresponding ARMDP $M_\phi$ based on a pCTL sentence $\phi$, checking $\phi$ against $M_\phi$ is equivalent to checking $\phi$ against $M_b$, formally,
    \begin{align*}
        M_b\models\phi \Leftrightarrow M_\phi\models\phi
    \end{align*}
\end{theorem}

Theorem~\ref{theorem1} is proven by using probabilistic bisimulation. Probabilistic bisimulation is a standard model checking technique that compares two probabilistic transition systems~\citep{Baier:2008}. 
Given a pCTL formula $\phi$, if two states from different transition systems are \textit{probabilistic bisimilar}, then their behaviors are indistinguishable. Hence, checking $\phi$ in either system yields identical results. 
Theorem~\ref{theorem1} extends the theorem by~\citet{Belardinelli2011}[Theorem 2] for the probabilistic setting \footnote{More specifically, as compared to their work, this paper uses RMDPs instead of artifact systems, MDPs instead of Kripke structures, ground states instead of databases, and probabilistic bisimulation instead of bisimulation. }.

We prove Theorem~\ref{theorem1} in two steps.
First, we define indistinguishable states (Definition~\ref{def: indistinguishable}) and probabilistic bisimulation (Definition~\ref{def: probabilistic bisimulation}). 
Second, we show that an ARMDP and its underlying state-bounded MDP actually define a probabilistic bisimulation (Proposition~\ref{prop1} and \ref{prop2}).

\begin{define}
\label{def: indistinguishable}
Given the $b$-bounded MDP $M_b = \langle S_b, A_b, T_b\rangle$ of a RMDP $K=\langle \Sigma,\Delta\rangle$ and a pCTL sentence $\phi$, consider two ground states $s_1$ and $s_2$ in $S_b$ such that  $C_1=consts(s_1)\subset D$ and $C_2=consts(s_2)\subset D$. 
Let $C$ be the set of all $\phi$-relevant domain objects
$
C := consts(\phi)\cup consts(\Delta)\subseteq C_1\cap C_2
$.
The two states $s_1$ and $s_2$ are called \textbf{indistinguishable} \footnote{Indistinguishable states are also commonly called isomorphic states.} under $C$, if and only if $s_1$ and $s_2$ are renamings of one another under a bijection $f: C_1\setminus C\mapsto C_2\setminus C$. The bijection $f$ renames $\phi$-irrelevant domain objects. 
Formally, 
\begin{align*}
    s_1\sim_C s_2 \iff \; f(s_1) = s_2 
\end{align*}
We abuse the notation and let $f(s_1)$ be the state obtained by renaming constants in $s_1$. 
% an ARMDP $M_\phi = \langle S_\phi, \Sigma, \Delta, W\rangle$ of the
\end{define}

Definition~\ref{def: indistinguishable} has defined indistinguishable states under a pCTL formula $\phi$. 
Indistinguishable states are essentially renamings of one another and share the same properties with respect to a formula $\phi$, hence, they can be represented by one abstract state in an ARMDP. 
That is, an ARMDP state represents infinitely many indistinguishable states in the underlying MDP. 

\begin{example}
(Cont. Example~\ref{ex: ARMDP}) Consider a state-bounded MDP $M_b=\langle S_b, A_b, T_b\rangle$ with a state bound $b=2$, a pCTL formula $\phi = \mathtt{cl(a)}$ and two ground states
$s_3=\{\mathtt{cl(b), on(b, a)})\}\in S_b$ and $s_4=\{\mathtt{cl(c), on(c, a)})\}\in S_b$. 
States $s_3$ and $s_4$ are variable renamings of each other thus indistinguishable under $\phi$.
They both express \textit{some block is on block $\mathtt{a}$} and can be represented by the abstract state $s_5=\{\mathtt{cl(Bl_1), on(Bl_1,a)}\}$.
\end{example}

Now we must define \textit{probabilistic bisimulation}. 

\begin{define} 
\label{def: probabilistic bisimulation}(cf.~\citet{Baier:2008})
Consider a $b$-bounded MDP $M_b=\langle S_b,A_b,T_b\rangle$, a pCTL formula $\phi$ and the finite set $C$ of $\phi$-relevant constants, 
$\mathcal{R}\subseteq S_b\times S_b$ is a \textbf{probabilistic bisimulation} if for any pair of states $\langle s_1,s_2\rangle\in \mathcal{R}$:
\begin{enumerate}
    \item $s_1$ and $s_2$ are indistinguishable under $C$, i.e. $s_1\sim_C s_2$
    \item $s_1$ and $s_2$ have the identical probabilities of going to any other distinguishable states, i.e. $\forall a\in A: T(s_1,a)(s) = T(s_2,a)(s)$ for each $s\in S/\mathcal{R}$
\end{enumerate}
States $s_1$ and $s_2$ are called probabilistic bisimilar.
\end{define}

To prove Theorem~\ref{theorem1}, we must prove that $M_\phi$ and $M_b$ indeed form a probabilistic bisimulation. 
Propositions~\ref{prop1} and \ref{prop2} will together prove Theorem~\ref{theorem1}, i.e. checking an ARMDP is equivalent to checking its underlying state-bounded MDP.

\begin{proposition}
\label{prop1}
For an ARMDP $M_\phi = \langle S_\phi,\Sigma,\Delta, W\rangle$ based on a $b$-bounded MDP $M_b = \langle S_b, A,_b T_b\rangle$ and a pCTL sentence $\phi$, if two states are indistinguishable, then they either both can fire a transition rule or both cannot fire a transition rule. 
Formally, for any transition rule of the form
    $$H_i \xleftarrow{p_i:\alpha} B$$
if $s_1\sim_C s_2$, then 
\begin{align*}
s_1 \preceq_{\theta_1} B \iff s_2 \preceq_{\theta_2} B
\end{align*}
for two indistinguishable substitutions $\theta_1$ and $\theta_2$. 
\end{proposition}

\begin{proposition}
\label{prop2}
For an ARMDP $M_\phi = \langle S_\phi,\Sigma,\Delta, W\rangle$ based on a $b$-bounded MDP $M_b = \langle S_b, A_b, T_b\rangle$ and a pCTL sentence $\phi$, if two states are indistinguishable, then they have the identical probabilities of going to any other distinguishable states. 
Formally, for any transition rule $\delta\in\Delta$ of the form
    $$\{H_1 \xleftarrow{p_1:\alpha} B, ..., H_n \xleftarrow{p_n:\alpha} B\}$$
if $s_1\sim_C s_2$, then they define two sets of ground transition rules $T(s_1,\alpha\theta_1)$ and $T(s_2,\alpha\theta_2)$ that share the same transition probabilities.
\begin{align*}
    T(s_1,\alpha\theta_1) := \{ h_{1,i} \xleftarrow{p_i:\alpha\theta_1} s_1 | 
&H_i \xleftarrow{p:\alpha} B\in \delta, s_1 \preceq_{\theta_1} B,\\
&h_{1,i} = (s_1\backslash B\theta_1)\cup H_i\theta_1
\} 
\end{align*}
\begin{align*}
    T(s_2,\alpha\theta_2) := \{ h_{2,i} \xleftarrow{p_i:\alpha\theta_2} s_2 | 
&H_i \xleftarrow{p:\alpha} B\in \delta, s_2 \preceq_{\theta_2} B,\\
&h_{2,i} = (s_2\backslash B\theta_2)\cup H_i\theta_2
\} 
\end{align*}
\end{proposition}

Propositions~\ref{prop1} and \ref{prop2} have shown that the indistinguishable relation $\sim_{C}$ between an ARMDP and its underlying state-bounded MDP is a probabilistic bisimulation. $\sim_{C}$ is an equivalence relation such that two probabilistic bisimilar states exhibit identical behavior. In other words, the ARMDP and the underlying MDP denote a mutual, step-wise simulation of indistinguishable states.  
With a similar reasoning as in~\citep{Belardinelli2011}, we conclude that the model checking problem for infinite, state-bounded MDPs is decidable as it can be done on its corresponding finite ARMDP.

\subsection{PCTL-REBEL for Infinite MDPs}
\label{sec: pCTL REBEL infinite MDPs}
PCTL-REBEL can handle an infinite, state-bounded RMDP by reasoning about its finite abstraction. More formally, given an infinite RMDP with a step bound and a pCTL formula, a finite ARMDP can be constructed and naturally checked by pCTL-REBEL. 
In fact, as an ARMDP is just like an RMDP that concerns at most $b$ objects in a state (see Section~\ref{sec: abstract MDPs}), pCTL-REBEL requires only one adaption to restrict the state size.
That is, all states that have more than $b$ objects must be eliminated. The number of objects in a state bound is simply obtained by counting all variables and constants as OI-subsumption is imposed.
The adaption of maintaining the state bound is implemented in \texttt{Regression} (Algorithm~\ref{algo:Regression}).

PCTL-REBEL for an infinite, state-bounded MDP is lifted and complete, just as like pCTL-REBEL for a finite RMDP (see Section~\ref{sec: properties of pCTL-REBEL}). We discuss properties of pCTL-REBEL for state-bounded MDPs in detail.
\begin{description}
\item[\textbf{Lifted}] PCTL-REBEL checks a state-bounded, infinite RMDP that has a finite abstraction (namely, an ARMDP) at a lifted level. Specifically, pCTL-REBEL operates on the ARMDP that is bisimilar to the underlying ground MDP. Checking an ARMDP is the same as checking an RMDP, except that the state bound must be maintained.

\item[\textbf{Sound for step-bounded pCTL formulae}] PCTL-REBEL is sound for step-bounded pCTL formulae in an ARMDP.
This is a consequence of Theorem~\ref{theorem1} that shows checking an ARMDP yields exactly the same results as checking its underlying MDP.
PCTL-REBEL is not sound for indefinite-horizon formulae but it achieves precise approximation in practice. 

\item[\textbf{Complete}] PCTL-REBEL is complete for checking ARMDPs as all states are assigned a probability. PCTL-REBEL captures the entire state space by using an ordered set of relational $V^p$-rules. 
\end{description}

\section{Experiments}
\label{sec:experiments}
We aim to answer the following questions in this section. Q1 and Q2 focus on the benefits of pCTL-REBEL, and Q3-Q5 focus on the limitations of pCTL-REBEL.
\begin{itemize}
    \item[\textbf{Q1}] What formulae can pCTL-REBEL check in practice?
    \item[\textbf{Q2}] How does pCTL-REBEL compare with state-of-the-art model checkers?
    \item[\textbf{Q3}] How well does pCTL-REBEL handle indefinite-horizon formulae?
    \item[\textbf{Q4}] How well does pCTL-REBEL handle a complex relational transition function?
    \item[\textbf{Q5}] What are the computational costs of different pCTL operators?
\end{itemize}

We implemented and validated an unoptimized pCTL-REBEL research prototype using SWI-Prolog 8.0.2, with the \textit{constraint handling rules} library. Experiments were run on a 2.4 GHz Intel i5 processor. 
% The execution time was estimated by SWI-Prolog's built-in predicate $\mathtt{statistics(run\_time, \cdot)}$. 
We use the blocks world dataset \footnote{https://qcomp.org/benchmarks/\#blocksworld} and the box world dataset \footnote{https://qcomp.org/benchmarks/\#boxworld}.
We compare pCTL-REBEL with the state-of-the-art model checkers PRISM~\citep{Prism:2011} and STORM~\citep{Storm:2017} \footnote{The experiments will be made public once the paper is accepted}. 
% and ePMC~\citep{JaniPrismConverter}. 
We set a time-out of 1800 seconds for all model checkers. PCTL-REBEL's state bound is sometimes referred as \textit{number of blocks/cities} so as to provide a direct comparison with PRISM and STORM that operate on ground models.

The blocks world and the box world datasets originate from IPPC-2008 \footnote{http://ippc-2008.loria.fr/wiki/index.php/Results.html\#Fully\_Observable\_Non-Deterministic\_.28FOND.29\_track\_2} and are originally specified in the ppddl language. STORM and PRISM operate on the converted prism-format files \footnote{The ppddl models are first converted to the jani format by the ppddl2jani tool (attached in the datasets), then converted to the prism format by ePMC~\citep{JaniPrismConverter}. }, and pCTL-REBEL operates on the converted prolog-format files \footnote{The ppddl models are translated to prolog models as PCTL-REBEL is implemented in prolog. }. 
We discard the reward structure in the model. 
The blocks world dataset is simplified to have the relations $\mathtt{on/2, clear/1}$ 
and the action $\mathtt{move/3}$ that has a success probability of 0.9.
The box world dataset has the relations $\mathtt{bin/2, on/2, tin/2, can\mhyphen drive/2}$
and the actions $\mathtt{drive/3, load/2, unload/2}$. 
In the box world, $\mathtt{bin(B, C)}$ expresses a box is in a city, $\mathtt{on(B, T)}$ expresses a box is on a truck, and $\mathtt{tin(T, C)}$ expresses a truck is in a city. An atom $\mathtt{can\mhyphen drive(C1,C2)}$ expresses a road that directly connects $\mathtt{C1}$ and $\mathtt{C2}$. 
A box can be loaded on ($\mathtt{load(B, T)}$) or unloaded from a truck ($\mathtt{unload(B, T)}$).
A truck can travel from a city to another ($\mathtt{drive(T, C1, C2)}$). The $\mathtt{load/2}$ and $\mathtt{unload/2}$ actions succeed with probability 0.9. The $\mathtt{drive/2}$ action succeeds with probability 0.8.

\subsection*{Q1: What formulae can pCTL-REBEL check in practice?} 
Although pCTL-REBEL can check any formulae of the pCTL language as discussed in Section~\ref{sec:pCTL}, in practice, some formulae are more costly than others. This experiment aims at evaluating the computational costs of different pCTL formulae.
Table~\ref{tab: showcases} includes formulae ranging from classic planning properties, i.e. reachability properties~\citep{Kersting:2004, Boutilier:2001, Sanner:2009}, to more complex, nested pCTL properties, e.g. the following $\phi_{\mathtt{nested}}(i,j)$ formula. Recall that $\phi_{\mathtt{nested}}(4,2)$ was discussed in Example~\ref{ex: nested formula}, and its parse tree is in Figure~\ref{fig:parse tree}. Table~\ref{tab: showcases} also includes a formula that has an indefinite horizon, namely Property 5. 
\begin{align*}
\phi_{\mathtt{nested}}(\textcolor{blue}{i}, \textcolor{blue}{j}) = 
\mathtt{P_{\geq 0.5}[cl(a)U^{\leq \textcolor{blue}{i}} (on(a,b)}
\mathtt{\wedge P_{\geq 0.8} [P_{\geq 0.9}[X\;cl(e)]U^{\leq \textcolor{blue}{j}}on(c,d) ])]}
\end{align*}

\begin{figure}
\floatbox[{\capbeside\thisfloatsetup{capbesideposition={right,top},capbesidewidth=0.4\textwidth}}]{figure}[\FBwidth]
{\caption{The abstracts states that satisfy $\mathtt{P_{\geq 0.5}[F^{\leq 10} on(a,b)]}$. Any state subsumed by $\{\mathtt{on(a,b)}\}$ is a goal state thus has a value 1. The states that are further away from the upper-left corner require more successful actions to reach $\{\mathtt{on(a,b)}\}$ thus have a smaller probability. }
 \label{fig: blocksworld 10 steps}}
{\includegraphics[width=0.5\textwidth]{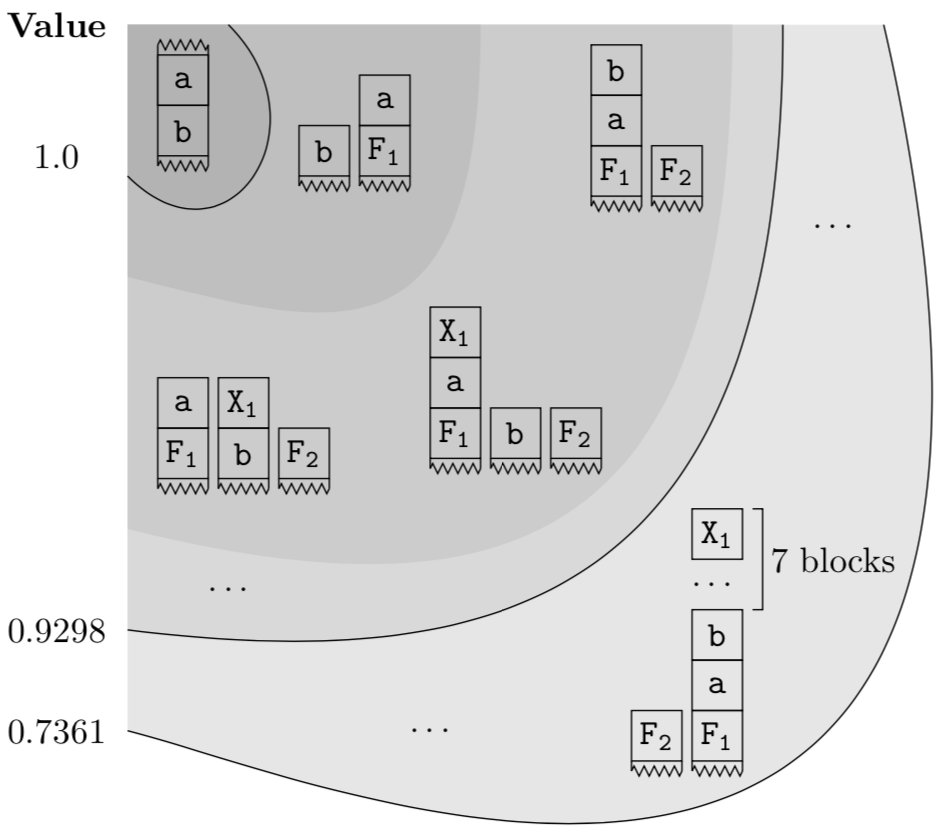}}
\end{figure}

\begin{figure}
\floatbox[{\capbeside\thisfloatsetup{capbesideposition={right,top},capbesidewidth=0.3\textwidth}}]{figure}[\FBwidth]
{\caption{Similar to Figure~\ref{fig: blocksworld 10 steps}, this illustrates the abstract states that satisfy $\phi_{\mathtt{nested}}(4,2)$ in the blocks world. }
 \label{fig:nested formula exp}}
{\includegraphics[width=0.6\textwidth]{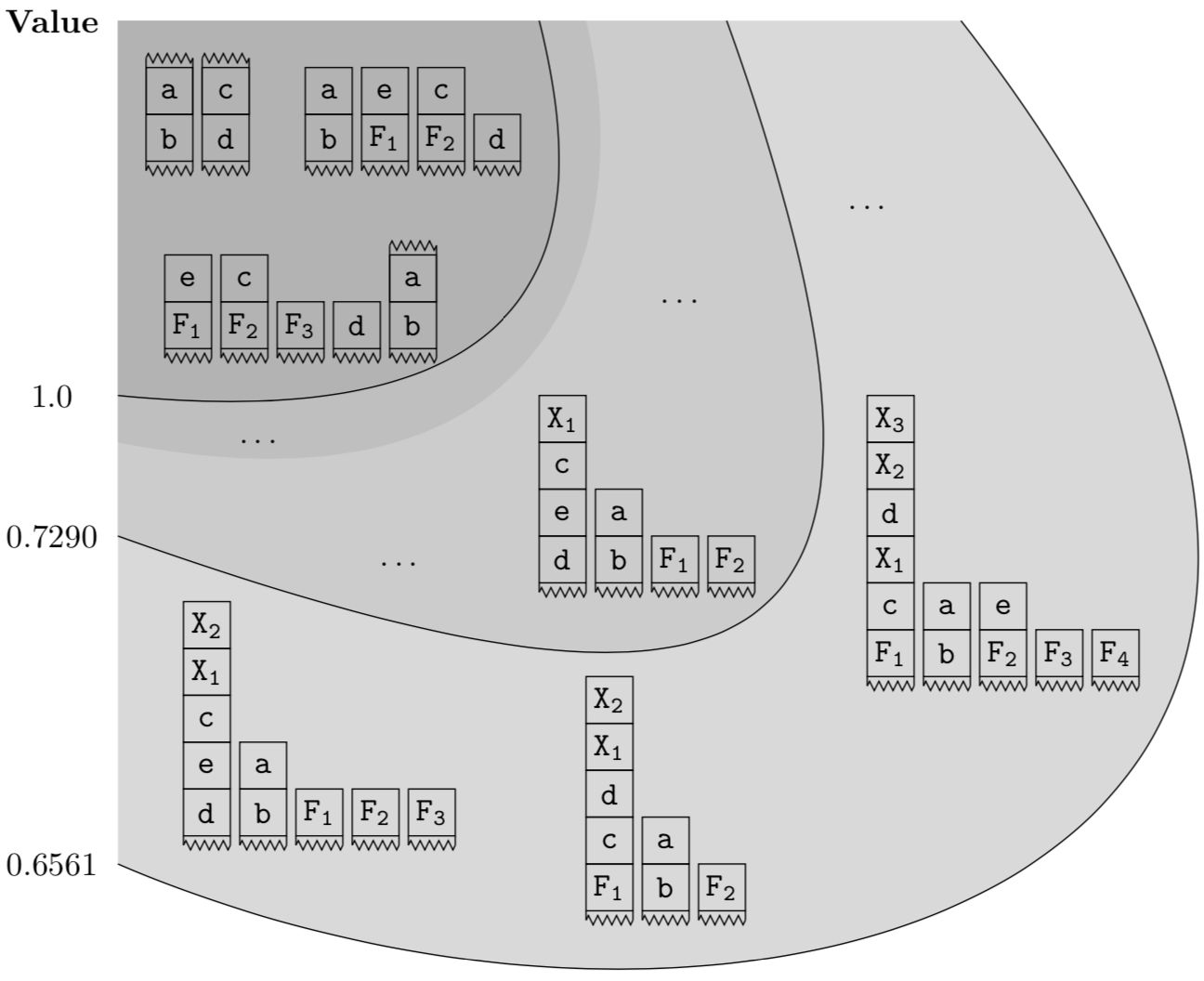}}
\end{figure}

We now discuss Table~\ref{tab: showcases} in detail. 
First, all formulae in Table~\ref{tab: showcases} are checked against models that have \textit{infinitely many} objects, which is possible only because pCTL-REBEL uses lifted inference. 
Second, pCTL-REBEL can handle classic planning tasks by formulating them as reachability formulae (Property 1,2,5, Table~\ref{tab: showcases}) or constrained reachability formulae (Property 3,6, Table~\ref{tab: showcases}). For example, Figure~\ref{fig: blocksworld 10 steps} visualizes the solution to Property 2 in Table~\ref{tab: showcases}. 
Third, pCTL-REBEL can check, in an infinite model, formulae that have an infinite horizon (Property 5, Table~\ref{tab: showcases}). This property cannot be evaluated by an explicit-state model checker as state space is infinite. 
Finally, pCTL-REBEL can handle nested formulae such as $\phi_{\mathtt{nested}}(i,j)$ (Property 4, Table~\ref{tab: showcases}). Under a time-out of 1800 seconds, pCTL-REBEL can compute $\phi_{\mathtt{nested}}(3,1)$, which is cheaper than computing $\phi_{\mathtt{nested}}(4,2)$. Nonetheless, Figure~\ref{fig:nested formula exp} visualizes the results of $\phi_{\mathtt{nested}}(4,2)$ by relaxing the time-out. More discussion about nested formulae will be in Q5.

\begin{table}
\centering
    \begin{tabular}{c|c|c|c}
    domain & property  & formula & runtime(sec) \\ \hline \hline
    blocks & \begin{tabular}[c]{@{}l@{}}bounded reachability\\ with det. actions\end{tabular} & $\mathtt{P_{\geq0.5}[F^{\leq 10} on(a,b)]}$  & 11 \footnote{\citet{Kersting:2004} reported a runtime of roughly 10 minutes for this task} \\ \hline
    blocks & bounded reachability & $\mathtt{P_{\geq0.5}[F^{\leq 10} on(a,b)]}$  & 226 \\ \hline
    blocks & \begin{tabular}[c]{@{}l@{}}bounded\\ constrained reachability\end{tabular}  & $\mathtt{P_{\geq0.5}[on(c,d) U^{\leq 5} on(a,b)]}$  & 7.1 \\ \hline
    blocks &  nested   & $\phi_{\mathtt{nested}}(3,1)$  & 1554 \\ \hline
    box & unbounded reachability & $\mathtt{P_{\geq0.5}[F\; bin(b1,paris)]}$ & 0.7 \\ \hline
    box & \begin{tabular}[c]{@{}l@{}}bounded \\constrained reachability\end{tabular} & $\mathtt{P_{\geq0.5}[tin(t1,paris) U^{\leq 5} bin(b1,paris)]}$  & 0.365
    \end{tabular}
    \caption{PCTL-REBEL can perform a range of pCTL properties in the \textit{infinite} blocks and box worlds under a time-out of 1800 seconds. }
    \label{tab: showcases}
\end{table}

\subsection*{Q2: How does pCTL-REBEL compare with state-of-the-art model checkers?}
State-of-the-art model checkers STORM and PRISM do not operate on an RMDP as in Q1. They require the RMDP to be grounded, which leads to a state explosion. 
This experiment evaluates how well STORM and PRISM handle such state explosions in practice.
In particular, we show that pCTL-REBEL, by lifting, performs better in mitigating state explosions in relational domains. 
While Q1 assumes that the given RMDP has an infinitely large domain, in order to compare pCTL-REBEL with STORM and PRISM, this section considers finite domains such that explicit-state models can be generated. Furthermore, we increment the domain size by 1 to track the trends in computation time for all model checkers.

% use jani public datasets: https://qcomp.org/benchmarks/

\begin{figure}
\floatbox[{\capbeside\thisfloatsetup{capbesideposition={right,top},capbesidewidth=0.3\textwidth}}]{figure}[\FBwidth]
{\caption{The blocks world stats. 
    When the number of blocks grows, the number of ground states and the number of nondeterministic transitions explode more than the minimum number of abstract states.}
 \label{fig: statenumber}}
{\includegraphics[width=0.65\textwidth]{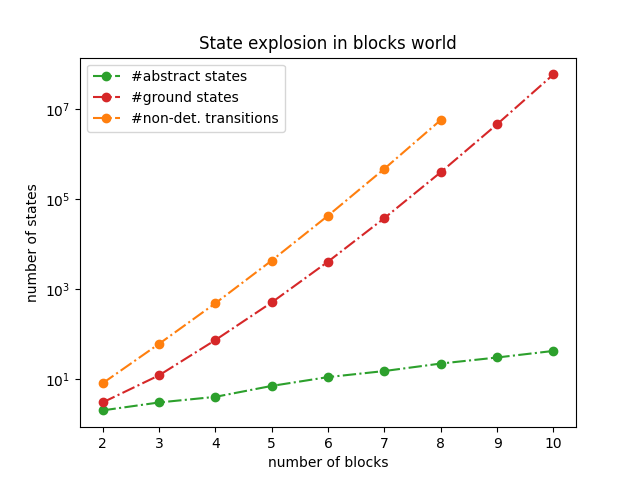}}
\end{figure}

The state explosion problem is illustrated in Figure~\ref{fig: statenumber}, showing the number of ground states~\citep{slaney2001blocksworld} and the corresponding ground transitions \footnote{The number of ground transitions is obtained from the STORM model checker. } grow much faster than the minimum number of abstract states \footnote{The minimum number of abstract states is the number of integer partitions that capture all relational structures. For example, for three blocks, there are three abstract states $\mathtt{\{cl(A), cl(B), cl(C)\}}$, $\mathtt{\{cl(A), cl(B), on(B,C)\}}$ and $\mathtt{\{cl(A), on(A,B), on(B,C)\}}$. }. Even though all numbers grow exponentially, the minimum number of abstract states is the most resistant to the growth of the domain size.

% \begin{table}[ht]
%     \centering
%     \begin{tabular}{c||c|c|c}
%     % blocks & \begin{tabular}[c]{@{}c@{}}abstract\\ 
%     % states\end{tabular} & \begin{tabular}[c]{@{}c@{}}ground\\ 
%     % states\end{tabular} & \begin{tabular}[c]{@{}c@{}}non-det. \\ \hline
%     \#blocks & \#abstract states & \#ground states & \#non-det. ground transitions \\ \hline
%     2  & 2  & 3          & 8       \\ \hline
%     3  & 3  & 13         & 60      \\ \hline
%     4  & 5  & 73         & 480     \\ \hline
%     5  & 7  & 501        & 4 280   \\ \hline
%     6  & 11 & 4 051      & 42 600  \\ \hline
%     7  & 15 & 37 633     & 470 148 \\ \hline
%     8  & 22 & 394 353    & 5 707 520\\ \hline
%     9  & 30 & 4 596 553  &  -       \\ \hline
%     10 & 42 & 58 941 091 &  -       \\ 
%     \end{tabular}
%     \caption{The blocks world stats. 
%     When the block number grows, the number of ground states and the number of nondeterministic transitions explode more than the minimum number of possible relational structures. }
%     \label{tab: statenumber}
% \end{table}{}

\begin{figure}
\centering
\includegraphics[width=.7\textwidth]{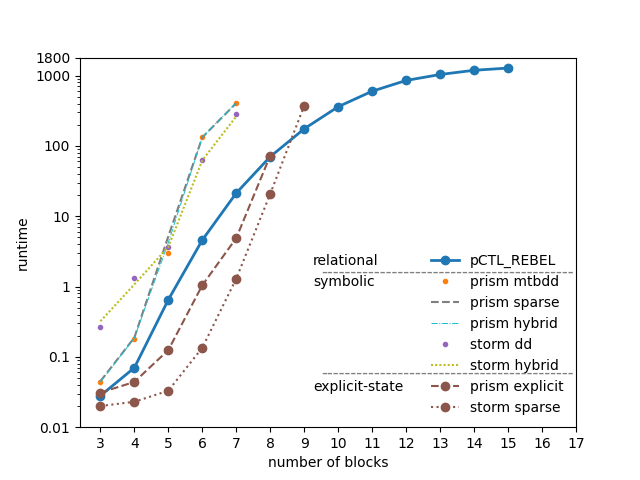}
\captionsetup{width=.7\linewidth}
\caption{The property $\mathtt{P_{\geq 0.5}[F^{\leq 10} on(a,b)]}$ of is checked by three model checkers. With a time-out of 1800 seconds, PRISM can handle at most 8 blocks with the explicit engine, STORM can handle at most 9 blocks with the sparse engine, and pCTL-REBEL can handle 15 blocks. }
 \label{fig: scalability}
\end{figure}

We compare the scalability of the three model checkers (including 4 PRISM engines and 3 STORM engines) in the blocks world.  
We could essentially select any pCTL property for the comparison. Nonetheless, for simplicity, we use a simple reachability property $\mathtt{P_{\geq 0.5}[F^{\leq 10} on(a,b)]}$ (Property 2, Table~\ref{tab: showcases}). 
We measure how the runtime grows with respect to the domain size, shown in Figure~\ref{fig: scalability}. The results show that pCTL-REBEL is more scalable than any PRISM or STORM engine in relational domains. Under a time-out of 1800 seconds, pCTL-REBEL can handle 15 blocks (6.6e13 ground states) whereas PRISM can handle at most 8 blocks (4.0e5 ground states) and STORM can handle at most 9 blocks (4.6e6 ground states).

We analyze Figure~\ref{fig: scalability} to compare lifting with other optimization techniques. Observe that Figure~\ref{fig: scalability} categorizes all engines in three groups: (1) \textit{relational} model checking: pCTL-REBEL, (2) \textit{symbolic} model checking: PRISM's mtbdd, sparse, hybrid engines and STORM's dd, hybrid engines, and (3) \textit{explicit-state} model checking: PRISM's explicit engine and STORM's sparse engine. 
Compared to pCTL-REBEL, not only the explicit-state approaches but also the symbolic approaches are much less resilient to the growth of the domain size. 
Clearly, although symbolic model checking methods are known to be able to handle large domains, their ability of handling state explosions in relational domains is limited. 
Furthermore, unlike other engines, the runtime of pCTL-REBEL saturates when the domain grows to a certain extent. This phenomenon is a consequence of lifted inference, reflecting that the number of relational structures will eventually saturate.

\subsection*{Q3: How well does pCTL-REBEL handle indefinite-horizon formulae?}
PCTL-REBEL suffers from runtime explosions when checking a formula that has an indefinite horizon. In specific, the runtime grows exponentially with respect to the horizon. This experiment aims at analyzing the cause of the runtime explosion.
We use the reachability property $\mathtt{P_{\geq0.5}[F\; on(a,b)]}$ that has an indefinite horizon.

To analyze the cause of the runtime explosion, we split the \textit{iterations} of checking an indefinite-horizon formula into two stages: (1) the \textit{state recognition stage} and (2) the \textit{value convergence stage}. In the state recognition stage, pCTL-REBEL identifies new satisfying states in each iteration. This stage ends when no more new satisfying states are found. In the value convergence stage, pCTL-REBEL does not discover new states, and only updates probabilities of all identified states until convergence. Notice that these two stages are just for analyzing purposes as pCTL-REBEL conducts exactly the same computation in all iterations in both stages.

\begin{figure}[ht]
    \centering
	\includegraphics[width=\textwidth]{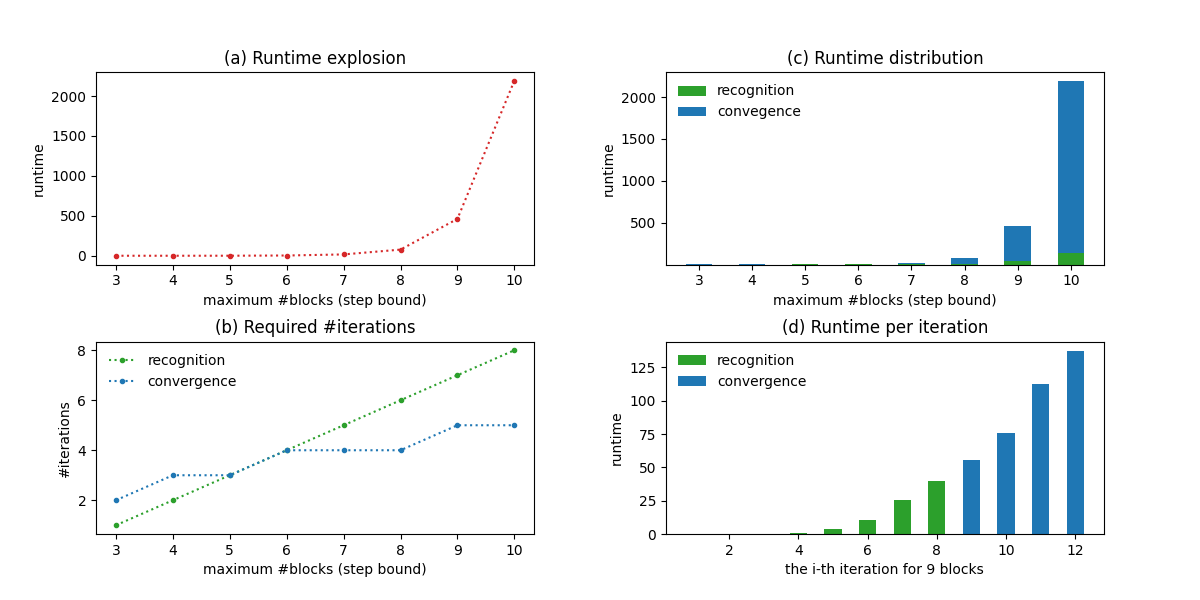}
	\caption{The results obtained from checking the reachability property $\mathtt{P_{\geq0.5}[F\; on(a,b)]}$ in the blocks world. 
	(a) The runtime increases exponentially to the number of required iterations. Most of the runtime is taken by the value convergence stage (b) The later iterations are more expensive than the earlier iterations.  }
	\label{fig:block world limit}
\end{figure}

We show the runtime growth of $\mathtt{P_{\geq0.5}[F\; on(a,b)]}$ in Figure~\ref{fig:block world limit} with respect to the domain size. 
Figure~\ref{fig:block world limit}(a) shows the runtime explodes with respect to \textit{the number of required iterations}, and illustrates that most of the runtime is taken by the value convergence stage. Moreover, Figure~\ref{fig:block world limit}(b) shows that within one checking process, later iterations are more expensive than earlier iterations. This means that although the job of the value convergence state is seemingly easier than the state recognition stage, in practice, it takes more time. This extra computation time comes from pCTL-REBEL's attempts at identifying new states in the value convergence stage. 

In summary, pCTL-REBEL suffers from a runtime explosion when handling indefinite formulae. The main cause is the redundant computation conducted in the value convergence stage. That is, pCTL-REBEL performs the same computation without being aware of the two-stage nature of the model checking process.
This problem can be mitigated by optimizing the algorithm so as to reduce the time taken by the second stage. The optimization should benefit all indefinite-horizon formulae.

\subsection*{Q4: How well does pCTL-REBEL handle a complex relational transition function?}

The use of abstract actions is key to the scalability of RMDPs as it allows for reasoning about a large group of ground actions as a whole. Roughly speaking, the more ground actions an abstract action captures, the more scalable pCTL-REBEL can be. For example, the blocks world domain has a simple $\mathtt{move}$ abstract action (see Figure~\ref{fig:abstract transition}) that captures countless ground actions. Therefore, pCTL-REBEL is very scalable to the growth of the domain size.  
However, it is common to have constraints on the action, which significantly harms scalability. 
An action constraint harms scalability by prohibiting certain substitutions for an abstract action. In other words, it forces the abstract action to depend on the true identity of some objects. Action constraints break relational structures and result in  non-symmetric transition functions. The resulting transition function consists of more rules and cannot be represented as compactly anymore. We call this constrained transition functions \textit{relationally} complex.

The following example illustrates action constraints. Consider a box world domain where the action $\mathtt{drive(T, C1, C2)}$ is constrained by the underlying road map below \footnote{The road networks are automatically generated by the box world dataset. }. A truck $\mathtt{T}$ can drive from $\mathtt{C1}$ to $\mathtt{C2}$ only if $\mathtt{can\mhyphen drive(C1,C2)}$ exists. 
\begin{multicols}{3}
\begin{verbatim}
can-drive(city0,city2).
can-drive(city0,city1).
can-drive(city0,city6).
can-drive(city1,city0).
can-drive(city1,city6).
can-drive(city1,city3).
can-drive(city1,city4).
can-drive(city1,city2).
can-drive(city1,city5).
can-drive(city2,city0).
can-drive(city2,city1).
can-drive(city2,city6).
can-drive(city3,city1).
can-drive(city3,city6).
can-drive(city3,city4).
can-drive(city3,city5).
can-drive(city4,city1).
can-drive(city4,city3).
can-drive(city4,city6).
can-drive(city4,city5).
can-drive(city5,city4).
can-drive(city5,city1).
can-drive(city5,city3).
can-drive(city6,city0).
can-drive(city6,city1).
can-drive(city6,city2).
can-drive(city6,city3).
can-drive(city6,city4).
\end{verbatim}
\end{multicols}

\begin{figure}
\centering
\includegraphics[width=.7\textwidth]{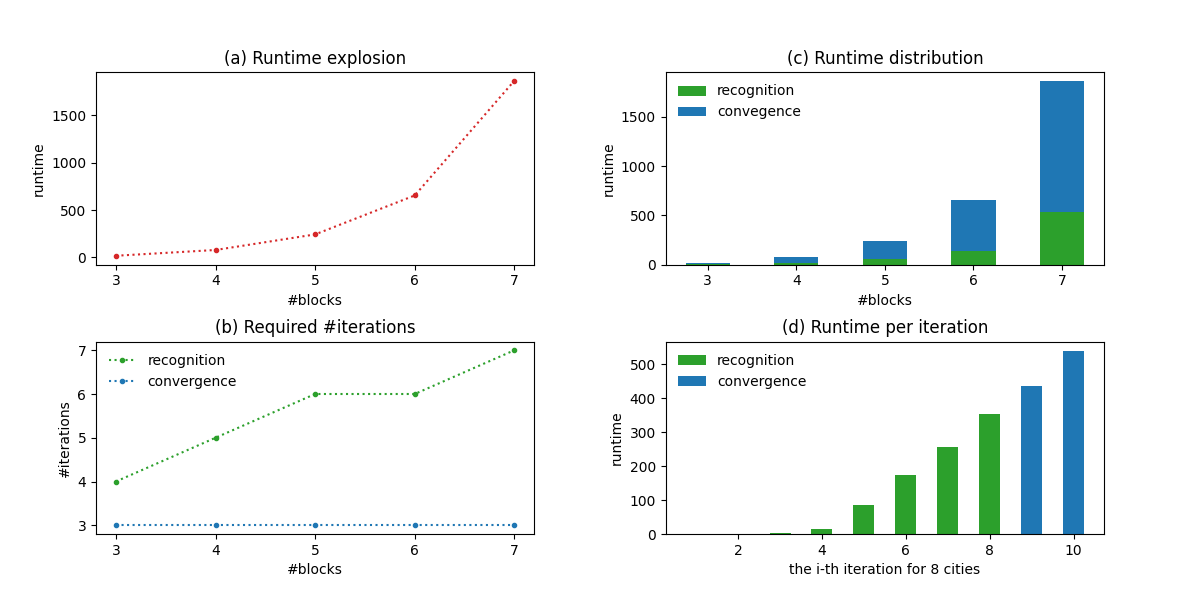}
\captionsetup{width=.7\linewidth}
\caption{The results obtained from checking the reachability property $\mathtt{P_{\geq0.5}[F\; bin(b1,city0)]}$ in the box world. The runtime increases exponentially when the transition structure becomes more complex. }
 \label{fig:box world limit}
\end{figure}

This experiment examines how well pCTL-REBEL handles complex relational transition functions. We use the reachability property $\phi = $ $\mathtt{P_{\geq0.5}[F\; bin(b1,city0)]}$ and automatically generated road maps. Figure~\ref{fig:box world limit} shows the results that pCTL-REBEL can handle a road map containing at most 8 cities under a time-out of 1800 seconds. The runtime increases exponentially with respect to the size of the road network. Recall that without a road network, it takes only 0.7 seconds to check $\phi$ (see Property 5, Table~\ref{tab: showcases}).
Since the road network makes the model dynamics more complex, more transition rules are required, resulting in runtime explosions. In short, pCTL-REBEL works the best when the dynamics of the objects is simple, e.g. in the blocks world or in the box world without a road map.

% \begin{figure}[ht]
% \floatbox[{\capbeside\thisfloatsetup{capbesideposition={right,top},capbesidewidth=0.4\textwidth}}]{figure}[\FBwidth]
% {\caption{The results obtained from checking the reachability property $\mathtt{P_{\geq0.5}[F\; bin(b1,city0)]}$ in the box world. The runtime increases exponentially when the transition structure becomes more complex. }
%  \label{fig:box world limit}}
% {\includegraphics[width=0.5\textwidth]{boxworld_limit.png}}
% \end{figure}

\subsection*{Q5: What are the computational costs of different pCTL operators?}
\label{exp: pctl operation cost}

It is crucial to know what kind of formulae pCTL-REBEL can handle efficiently when designing properties for a model. This experiment aims at giving insight into how different factors affect pCTL-REBEL's performance.

Some path operators are more costly than others. 
e.g. the $\mathtt{X}$ operator is the cheapest since it requires only one iteration. The reachability operator $\mathtt{F}$ is generally more expensive than the constrained reachability operator $\mathtt{U}$ since $\mathtt{U}$ is a stricter operator that prunes out more states. For example, checking $\mathtt{P_{\leq 0.9} [F^{\leq 7}\;on(a,b)]}$ (18.494 seconds) is more costly than checking $\mathtt{P_{\leq 0.9} [on(c,d)\;U^{\leq 7}\;on(a,b)]}$ (0.733 seconds). A nested formula is more costly than a flat formula since it requires several recursive checking processes.

\begin{figure}
\centering
\includegraphics[width=.7\textwidth]{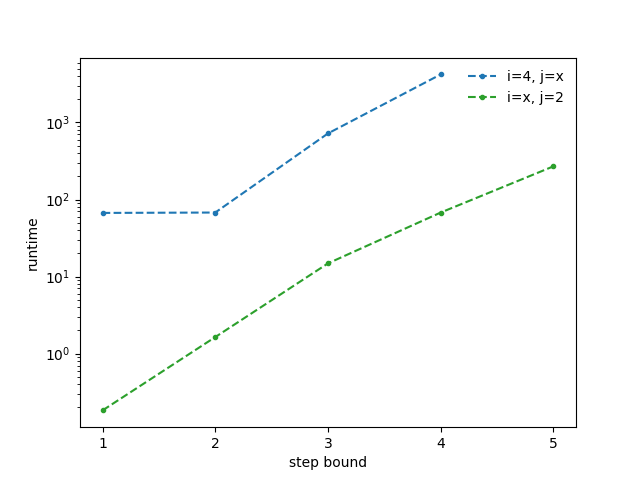}
\captionsetup{width=.7\linewidth}
\caption{The runtime of $\phi_{\mathtt{nested}}(i,j)$ for different values of i and j. The argument i is the step bound of the outer $\mathtt{U}$ operator, and j is the step bound of the inner $\mathtt{U}$ operator. The runtime grows exponentially when either step bound increases. However, the runtime is more sensitive to the inner step bound j. }
 \label{fig:stepbound}
\end{figure}

The formula structure also influences runtime. Specifically, changing an inner step bound has a larger impact than changing an outer step bound. Taking the formula $\phi_{\mathtt{nested}}(i,j)$ (defined in Q1) for example, the parameters $\mathtt{i}$ and $\mathtt{j}$ are both step bounds for $\mathtt{U}$ formulae, but $\mathtt{i}$ is attached to an outer subformula and $\mathtt{j}$ is attached to the inner subformula. Hence, the total runtime is more sensitive to the change in $\mathtt{j}$ than to the change in $\mathtt{i}$, as shown in Figure~\ref{fig:stepbound}. The cause of this difference is that the inner formulae are computed before the outer formulae. By changing an inner step bound, the number of satisfying states for the inner subformula changes, which directly influence the overhead of computing the outer subformulae. 

\section{Related work}
\label{sec:related work}

We have introduced pCTL-REBEL, a relational model checking technique that reasons at the relational level and mitigates the state explosion problem in relational MDPs.
PCTL-REBEL extends existing frameworks. First, pCTL-REBEL extends pCTL model checking with relational representations.
As far as the authors know, pCTL-REBEL is the first relational technique for pCTL model checking. Second, pCTL-REBEL extends the decidability results of infinite, state-bounded MDPs to the probabilistic setting.

To mitigate the state explosion problem in model checking, several \textit{grouping} techniques have been proposed. Different from our work, the following methods exploit various types of symmetries other than relational structures.
\textit{Symbolic model checking}~\citep{McMillan1993} is a form of model checking that symbolically represents the state space and the set of states in which a formula holds. 
Symbolic model checking is often implemented using BDDs that are Boolean formulae in canonical form, in which isomorphic subformulae are shared, to allow efficient operations on the sets of states that they represent. Symbolic model checking has been first used for model checking of plain non-probabilistic finite state systems~\citep{McMillan1993}, then later adapted to the probabilistic case in~\citet{tutorialPRISM}. Contrary to our algorithm, these symbolic methods only apply to finite state MDPs. Another method to mitigate the state explosion problem is \textit{abstraction-refinement}, which successively generates abstractions by partitioning the state space into \textit{regions}. Therefore, the model checking algorithms handle regions instead of individual states~\citep{AbstractionRefinement:2007,symbolicAbstractionRefinement:2008}. These techniques focus on grouping states into regions which have similar local properties but such regions are usually not exact as the states in a region can have different dynamics and future evaluations. 
Similarly,~\citet{Marthi2007} uses \textit{abstract MDP} to aggregate not only states but also actions by exploiting the temporal dependency of actions. 
In addition, \textit{game-based abstraction} techniques construct abstractions of MDPs by merging concrete transitions into abstract transitions~\citep{KATTENBELT20085}. Different from our work, for the abstractions, they calculate approximate upper and lower bounds instead of exact values.

Many studies integrate model checking techniques into other AI fields. 
In planning, authors compute the optimal policy at the first-order level using value iteration~\citep{Boutilier:2001,Sanner:2009,Kersting:2004}, which can be seen as a probabilistic reachability task, namely a special case of the general probabilistic model checking problem. PCTL-REBEL generalizes this special case with a temporal logic thus fits in the Planning as Model Checking paradigm~\citep{Giunchiglia:2000}. 
In robotics, model checking techniques are used for online motion planning~\citep{Lahijanian:2012,Maly:2013,He:2015} where the states are labeled with propositions. These propositions result in a multidimensional state space that grows exponentially with the domain size~\citep{He:2015}, which is the exact problem that this paper tackles. 
Generally, our work can be applied a wide range of frameworks that have a large model, particularly in relational domains.

There is a fruitful literature on theoretical results of infinite systems. Particularly, the state-boundedness condition for decidable verification has been defined in first-order mu-calculus~\citep{Hariri2012,Calvanese2016,DeGiacomo2012,DeGiacomo2016} and first-order CTL~\citep{Belardinelli2011,Belardinelli2012}. However, these studies focus on the non-probabilistic setting.
Our work contributes to extending the theoretical results to stochastic models by introducing the abstract relational MDP.
A limitation of the relational pCTL language is that it does not support quantification across conjunctions, which could be investigated as described in the work of~\citet{Belardinelli2013}.

\section{Towards Safe Reinforcement Learning}
\label{sec:discussion}

Recently, a converging interest has emerged about the pursuit of general dynamic systems that can autonomously \textit{reason} and \textit{learn} by interacting with the environment~\citep{degiacomo_ijcai19talk}. Since such systems require the ability of reasoning about first-order state representations and safety~\citep{amodei2016}, pCTL-REBEL fits in as a first-order model checking technique that is also particularly relevant in \textit{safe reinforcement learning}. 

Safe reinforcement learning is the process of learning a policy that maximizes the expected reward in domains where safety constraints must be respected. 
In particular, \textit{safe exploration}, i.e. guaranteeing an agent's safety in the exploration phase, is a nontrivial problem. 
Commonly used exploration strategies such as $\epsilon$-greedy and softmax sometimes select a \textit{random} action, which can result in catastrophic situations~\citep{amodei2016}. 
Existing research recognizes the importance of providing safety guarantees in reinforcement learning~\citep{Garcia2015,Pecka2014,Lahijanian:2012,teichteilkonigsbuch2012,sprauel2014,Giunchiglia:2000,Mason:2018,Hasanbeig2019,shielding2018,shielding2020}. However, most studies are based on techniques that require explicit state exploration whereas only few studies touch on safe exploration in relational domains~\citep{Driessens2004,MARTINEZ2017}
Furthermore, several studies have investigated augmenting reinforcement learning with model checking techniques, including preventing the agent from taking risky actions~\citep{shielding2018,shielding2020,Fulton2018}, synthesizing an initial safe \textit{partial} policy for learning~\citep{Leonetti2012}, learning a policy that maximizes the probability of satisfying a temporal formula~\citep{Hasanbeig2019}, and shaping the reward function based on a temporal formula~\citep{DeGiacomo2019}. 
These approaches can be augmented with our work for scalability in large relational domains. 

PCTL-REBEL can be used as a safe model-based reinforcement learning algorithm as it performs value iteration in a setting where goal states and safety constraints are formulated as pCTL formulae. 
More precisely, it derives the states in an RMDP that satisfy a given (relational) pCTL formula that encodes goal states and safety constraints.
Indeed, pCTL-REBEL tackles a special safe reinforcement learning task where the reward is 1 for goal states and 0 for all other states, and the discount factor is 1~\citep{Kersting:2004,yoon2012inductive,Otterlo:2004}. Formally, pCTL-REBEL in this paper 
\begin{align*}
V^p_{0}(s) &= 
\begin{cases}
0&, \forall s \not\in G\\
1&, \forall s \in G
\end{cases}\\
V^p_{t+1}(s) &= \max_{a} \sum_{s'} T(s,a,s')V^p_{t}(s')
\end{align*}
is a special case of the standard Bellman operator
\begin{align*}
V^p_{0}(s) &= 
\begin{cases}
0&, \forall s \not\in G\\
r&, \forall s \in G
\end{cases}\\
V^p_{t+1}(s) &= 
\max_{a} 
\sum_{s'} T(s,a,s')[R(s,a,s')+\gamma V^p_{t}(s')]
\end{align*}
where $r$ is a reward value in $\mathbb{R}$.

It is an interesting avenue for further research to investigate representing general reward functions within this framework. 
Furthermore, pCTL-REBEL can be naturally combined with relational reinforcement learning~\citep{dvzeroski2001relational} to achieve safe relational reinforcement learning as both frameworks use relational representations and work at an abstract level rather than at the ground level.

% Several Artificial Intelligence application areas show converging interests in

%~\citep{Brafman2019}: non-markovian rewards, temporal logics in planning

% The relational concept is recently used in deep reinforcement learning as well~\citep{zambaldi2018relational,janisch2020symbolic}.

%(2) Relation to planning (de giacomo slides in ijcai) and safe-RL (what does this contribute to?)\\
% \paragraph{Planning}
% In explainable AI,...

%~\citep{Brafman2019}: non-markovian rewards, temporal logics in planning

\section{Conclusions}
\label{sec:conclusion}

We have introduced a framework for lifted model checking in relational domains. To this aim, relational Markov Decision Processes have been integrated with model checking principles for pCTL. The result is a very expressive framework for model checking in probabilistic planning domains that are relational, that is, involve objects as well as the relations among them. 
The framework is lifted in that it does not require to first ground the relational MDP and then exhaustively check all possible paths, but rather works at a more abstract relational level where variables are only instantiated whenever needed. 
The resulting algorithm is based on the relational Bellman operator REBEL. 
It is quite complex but manages to rather compactly compress enormous spaces of ground states in a couple of 10s of rules. 
The algorithm has not yet been optimized, and one route for further work is to combine pCTL-REBEL with expressive first order decision diagrams~\citep{wang2008first} to gain efficiency and to further compress the abstract states. Another route for further work is to explore the use of pCTL-REBEL for reasoning and reinforcement learning in safety-critical contexts.

\begin{acknowledgements}
This work was supported by the FNRS-FWO joint programme under EOS No. 30992574.
It has also received funding from the Flemish Government under the “Onderzoeksprogramma Artificiële Intelligentie (AI) Vlaanderen” programme, 
the EU H2020 ICT48 project “TAILOR” under contract \#952215,
the Wallenberg AI, Autonomous Systems and Software Program (WASP) funded by the Knut and Alice Wallenberg Foundation,
and the KU Leuven Research fund. 
\end{acknowledgements}

% Authors must disclose all relationships or interests that 
% could have direct or potential influence or impart bias on 
% the work: 
%
% \section*{Conflict of interest}
%
% The authors declare that they have no conflict of interest.

% \bibliographystyle{authoryear}
% BibTeX users please use one of
\bibliographystyle{spbasic}      % basic style, author-year citations
\bibliography{ref}   % name your BibTeX data base

\end{document}